\documentclass[conference,compsoc]{IEEEtran}
\ifCLASSOPTIONcompsoc
  \usepackage[nocompress]{cite}
\else
  \usepackage{cite}
\fi
\ifCLASSINFOpdf
\else
\fi
\usepackage{tikz}
\usepackage{amsmath}
\usepackage{amssymb}
\usepackage{amsmath}
\usepackage{amsthm}
\usepackage{pifont}
\usepackage{booktabs}
\usepackage{url}
\theoremstyle{plain}
\newtheorem{definition}{Definition}
\usepackage{graphicx}
\usepackage{enumitem}
\usepackage{subcaption}  

\def\draftmode{1}

\def\paper{\draftmode}
\newcommand{\draft}[3]{\if\paper\draftmode\textcolor{#1}{#2}\else#3\fi}

\usepackage{todonotes} 

\newcommand{\takeawy}[1]{\todo[inline,caption={},color=gray!20]{{\it \textbf{Takeaway}:~}#1}}

\newcommand{\news}[1]{\todo[inline,caption={},color=green!15]{{\it \textbf{Why Object Unlearning? Why Scene Graphs?}~}#1}}

\begin{document}
%
\title{Targeted Therapy in Data Removal: Object Unlearning Based on Scene Graphs}

\author{
\IEEEauthorblockN{Chenhan Zhang\IEEEauthorrefmark{1}, Benjamin Zi Hao Zhao\IEEEauthorrefmark{1}, Hassan Asghar\IEEEauthorrefmark{1} and Dali Kaafar\IEEEauthorrefmark{1}}

\IEEEauthorblockA{
{\small \IEEEauthorrefmark{1}{School of Computing, Macquarie University, Macquarie Park, Australia}}
}}

\maketitle

\begin{abstract}

Users may inadvertently upload personally identifiable information (PII) to Machine Learning as a Service (MLaaS) providers. When users no longer want their PII on these services, regulations like GDPR and COPPA mandate a right to forget for these users. As such, these services seek efficient methods to remove the influence of specific data points. 
Thus the introduction of machine unlearning. 
Traditionally, unlearning is performed with the removal of entire data samples (sample unlearning) or whole features across the dataset (feature unlearning).
However, these approaches are not equipped to handle the more granular and challenging task of unlearning specific objects within a sample.
%
To address this gap, we propose a scene graph-based object unlearning framework. 
This framework utilizes scene graphs, rich in semantic representation, transparently translate unlearning requests into actionable steps.
The result, is the preservation of the overall semantic integrity of the generated image, bar the unlearned object.
Further, we manage high computational overheads with influence functions to approximate the unlearning process.
For validation, we evaluate the unlearned object's fidelity in outputs under the tasks of image reconstruction and image synthesis. 
Our proposed framework demonstrates improved object unlearning outcomes, with the preservation of unrequested samples in contrast to sample and feature learning methods.
This work addresses critical privacy issues by increasing the granularity of targeted machine unlearning through forgetting specific object-level details without sacrificing the utility of the whole data sample or dataset feature.
\end{abstract}


%
\IEEEpeerreviewmaketitle

\section{Introduction}



As machine learning models become increasingly integral to a range of personalized applications, from facial recognition to bespoke content generation, the protection of user privacy and the ability to comply with data removal requests have become paramount. The rise of Machine Learning as a Service (MLaaS) platforms has only intensified this need, as such platforms operate with large-scale, diverse datasets containing personal information.
With the implementation of stringent data privacy regulations, e.g., COPPA~\cite{COPPA1998} and GDPR~\cite{GDPR2018}, and recently e-Privacy~\cite{ePrivacy2019} and CCPA~\cite{CCPA2020},
users have the legal right to request the deletion of their personal data from these models. One way to entertain such requests is through \emph{machine unlearning}, a process where specific learned information is removed from a model without necessitating complete model retraining from scratch \cite{DBLP:conf/sp/CaoY15, nguyen2020variational, gupta2021adaptive, DBLP:conf/uss/ThudiJSP22,DBLP:conf/mm/ZhangBHX22}. Generally, these approaches either focus on removing entire samples or specific features across the training space. We argue that in many applications, these methods of machine unlearning are rather coarse-grained, and end up removing more information than necessary, thus adversely impacting the utility of the unlearned model. 


We illustrate our point through an example. Consider MLaaS for image generation or reconstruction. Privacy conscious users may request the removal of their personal data from these models. More specifically, the user wants the removal of his/her \emph{face} from any set of images used to train the model. 
Under existing unlearning methods, the service provider has two efficient approaches available to handle such a request. In the first instance, called \emph{sample unlearning} \cite{DBLP:conf/sp/CaoY15,SISA}, the service provider can remove all samples containing the user's face from the model. While this is good for images only containing the user's face, many images might be more complex containing other rich information such as cars or mountains in the background. These objects which may have no bearing on the user's privacy, yet still valuable to the model, would be removed as collateral damage. 

\vspace{3mm}
\news{
\\
\quad Consider a class reunion group photo uploaded to a social media platform, where others can tag you in the image. Now, suppose \textit{you, the privacy conscious individual wishes to have your face removed from the photo for privacy reasons, but the rest of the group has not made such a request}. How could only you be removed from any platform model using this data? 

\quad Further, consider the need to remove a boy from an image where he is wearing a cap, traditional segmentation methods might identify ``boy'' and ``cap'' as separate objects. Whilst the boy is removed, the cap is still strongly associated with said boy. How could object unlearning \textit{extend beyond simple object segmentation}?.
Enter, the scene graph, capable of capturing rich relationships, like ``boy has cap,'' allowing the unlearning process to ensure both the boy and his unique cap are removed together, or conversely ensure the presence of the hat.}
\vspace{3mm}

Alternatively, in the second approach, called \emph{feature unlearning} \cite{DBLP:conf/iccv/GandikotaMFB23, DBLP:conf/aaai/MoonC024}, the service provider can opt to erase all facial features from the model. However, this approach risks unintentionally removing facial data of other users who made no requests for erasure of their information, thereby diminishing the model's ability to accurately generate representations of faces of other users.

We can observe that both methods are \textit{coarse-grained} may unlearn much more than what is requested by the user, adversely affecting the model's overall performance. 
In this work, we develop machine unlearning techniques that work in a more granular level, in the sense that they unlearn parts of a sample while retaining utility both in terms of retaining information about other parts of the sample, and the impact on other samples with similar content. This application is most natural for images, which may contain multiple objects, only a subset of whom are requested to be removed. Another example is text-based data where only certain parts of the document are to be redacted.

\begin{table*}[ht]
    \caption{Five different types of machine unlearning. An illustration is given in the right part.}
    \hspace{-0.9cm}
    \begin{minipage}{0.5\textwidth} 
    \small
    \centering
    \resizebox{0.8\textwidth}{!}{%
    \begin{tabular}{l l l }
    \toprule
        \textbf{Unlearning Type} & \textbf{Unl. Request} $\mathbf{q}_{\mathrm{unl}}$   & \textbf{Unl. Granularity}  \\ \midrule
        
        Client Unlearning & $\mathcal{D}_{\mathrm{user}_i}$ &  $\blacksquare$ $\blacksquare$ $\blacksquare$ $\blacksquare$  \\ 

        Class Unlearning & $\forall I \in \Delta \mathcal{Y}$ &  $\blacksquare$ $\blacksquare$ $\blacksquare$  \\
        
        Sample Unlearning & $\forall I \in \Delta \mathcal{D}$ &  $\blacksquare$ $\blacksquare$  \\ 
        
        Feature Unlearning & $\forall \mathbf{o} \in \Delta \mathcal{C}$ &  $\blacksquare$ $\blacksquare$  \\ 

        Object Unlearning & $\mathbf{o} \in \Delta \mathcal{O}$ &  $\blacksquare$   \\ 
        \bottomrule
    \end{tabular}
    }
    \end{minipage}%
      \hspace{-0.5cm}
    \begin{minipage}{0.37\textwidth} 
    \centering
    \includegraphics[]{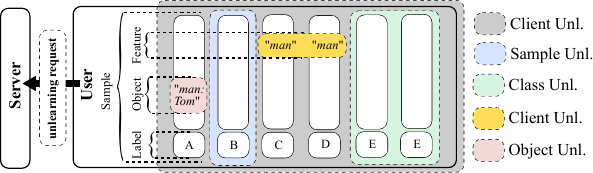} 
    \end{minipage}
    
    \label{tab:unl_type}
\end{table*}




We call this selective unlearning approach, \emph{object unlearning}, where object specifies individual entities within a sample, e.g., a physical object in an image or an entity in text. This unlearning approach is akin to \textit{targeted therapy} in medicine, where specific malignant cells are removed without damaging healthy tissue. 
A significant challenge of object unlearning is that objects do not exist in isolation; Instead there exists interwoven relationships between objects and their surrounding context that collectively contribute to the semantic coherence of the data sample. 
Unlearning a specific object while preserving the rest of the sample's content requires caution to ensure the removal of one element does not unintentionally create inconsistencies or degrade the model's understanding in the remaining structure or within other samples. As captured by the need to preserve the model’s performance and generalization capabilities. We contextualize the granularity of object unlearning in Table~\ref{tab:unl_type}.

To navigate these relational challenges, our approach for object unlearning leverages \emph{scene graphs}. Scene graphs provide a structured representation of an image by capturing objects, their attributes, and the relationships between them \cite{johnsonGF18}. This representation not only offers a high-level semantic understanding of visual content but also facilitates more nuanced and contextualized interpretations of scenes. Many studies have focused on generating scene graphs from images, and vice versa \cite{DBLP:journals/ijon/LiZZJDHSZSB24}. By leveraging the structure and semantics inherent in scene graphs, we can more precisely target objects for more effective, fine-grained unlearning techniques. 


Our contributions are summarized as follows:
\begin{itemize}[leftmargin=*, itemsep=0pt, topsep=0pt]
    \item This paper is the first to investigate the unlearning request of specific objects by MLaaS users. We identify the gap of fine-grained machine unlearning, one which allows the removal of specifically requested learned information while minimizing the impact on the model's overall utility. We formally propose the concept of object unlearning, which unlearns specific objects from an image.
    \item  To resolve the major challenge in object unlearning of disentangling interwoven objects, the assurance of one element's removal not unintentionally degrading the model in the remaining sample, we propose a scene graph-based object unlearning framework. 
    Scene graphs provide a direct and transparent means to translate unlearning requests into execution. 
    \item We comprehensively evaluate unlearning techniques developed in isolation for either sample or feature unlearning by adapting said techniques for all unlearning granularities. These techniques include influence functions, negative guidance, and masking techniques of patching and noise addition.
    \item Experimentation to validate the feasibility of unlearning objects covers tasks of both image reconstruction and image generation on benchmark datasets.
    \item The source code and artifacts of our proposed scene graph-based unlearning is released at \url{https://anonymous.4open.science/r/soul-24C8/}. 
\end{itemize}


\section{Related Work}\label{sec:rw_unl}
In this section, we first introduce recent studies in the field of machine unlearning, highlighting the relationships and differences between our work and existing techniques in graph unlearning and feature unlearning. Then, we also briefly discuss studies in scene graph and image manipulation, emphasizing their relevance to this work.
Machine unlearning is driven by individual privacy concerns and corresponding data privacy regulations such as GDPR \cite{GDPR} and CCPA \cite{CCPA}. 
A plethora of machine unlearning techniques have emerged in this trend. We shall introduce recent advancements in graph unlearning and feature unlearning techniques most closely related to our study.

\noindent\textbf{Graph Unlearning.}
Graph unlearning refers to the process of selectively removing the influence of specific nodes, edges, or subgraphs from a trained graph learning model (e.g., GNNs) \cite{grapheraser, gnndelete, wu2023gif, pan2023unlearning, wu2023certified, DBLP:conf/uss/WangH023}. 
For example, Chen \textit{et al.} \cite{grapheraser} extends SISA training for graph data with a graph partitioning technique to improve unlearning efficiency. 
Cheng \textit{et al.}'s \cite{gnndelete} learnable deletion operator extends GNNs for unlearning, to allow for unlearning without altering the GNN model's core weights.
Wu \textit{et al.} \cite{wu2023gif} utilize influence functions for rapid unlearning on graph nodes, edges, and node features.

However, existing graph unlearning methods primarily focus on learning tasks such as graph classification \cite{pan2023unlearning}, node classification \cite{wu2023gif} and link prediction \cite{gnndelete}, all of which are only applicable for graph-structured data. 
In contrast, our study addresses the unlearning of image data in its related tasks. We shall leverage graph unlearning techniques to achieve our object unlearning objective.

\noindent\textbf{Feature Unlearning.} 
Feature unlearning refers to removing a specific feature from a data sample while retaining the rest of the data sample \cite{DBLP:conf/satml/KongC23, DBLP:conf/ndss/WarneckePWR23, DBLP:journals/corr/abs-2303-17591,
DBLP:conf/iccv/GandikotaMFB23,
DBLP:conf/aaai/MoonC024, DBLP:journals/corr/abs-2404-03233, DBLP:journals/corr/abs-2402-11846}.
Guo \textit{et al.}'s seminal work \cite{guo2022efficient} proposed representation detachment to unlearn the specific attribute; However, only on supervised image classification tasks.
Several works have since considered feature unlearning on generative models \cite{DBLP:conf/ndss/WarneckePWR23,DBLP:conf/satml/KongC23, DBLP:journals/corr/abs-2303-17591, DBLP:conf/iccv/GandikotaMFB23, DBLP:conf/aaai/MoonC024}.
Warnecke \textit{et al.} \cite{DBLP:conf/ndss/WarneckePWR23} leverage influence functions to efficiently unlearn features and labels from generative language models. 
Kong and Chaudhuri \cite{DBLP:conf/satml/KongC23} propose a data augmentation-based algorithms for feature unlearning from pre-trained GANs.
Moon \textit{et al.} \cite{DBLP:conf/aaai/MoonC024} extracted latent representations corresponding to the target feature fur subsequent finetuning of the pre-trained generative model.
We note that in both \cite{DBLP:conf/satml/KongC23} and \cite{DBLP:conf/aaai/MoonC024}, there exists a need to collect specific images containing the target features, preparing such a specifically-crafted dataset for unlearning is labor-intensive. As \cite{DBLP:conf/aaai/MoonC024}, invited $13$ participants to manually annotate the data. 
Other research efforts focus on text-to-image diffusion models \cite{DBLP:conf/iccv/GandikotaMFB23, DBLP:journals/corr/abs-2303-17591, DBLP:journals/corr/abs-2402-11846}. Nevertheless, these methods are considerably restricted to text-to-image models based on cross-attention mechanisms, far from being a general technique. 

Further, a key distinction between feature unlearning and the proposed object unlearning is the former focuses on global image features. Whereas there exists instances where we only wish to unlearn the features of a specific object within the image. While Gandikota \textit{et al.} introduces the concept of erasing objects, their approach is coarse and erases an entire object class. That is to say, if there were three males (Males A, B, and C) in a generated image, all of them would be removed. In contrast, our proposed object unlearning achieves a more fine-grained unlearning based on scene graphs, allowing for the removal of only Male A while retaining Males B and C.

\noindent\textbf{Image Manipulation.}
Image manipulation or synthesis is the altering or transformation of images to achieve a desired effect or purpose \cite{DBLP:journals/pami/ZhanYWZLLKTX23}, for example, face swapping \cite{DBLP:conf/iccv/ShioharaYT23} and background replacement \cite{DBLP:conf/iccv/ChaiGWL23}. 
Many image manipulation methods can protect user privacy, with researchers using image synthesis techniques to conceal soft-biometric attributes of human faces while preserving the identity or keypoint matching regions of the facial image \cite{DBLP:conf/icit/WangWGZ21, DBLP:conf/aies/ZhuFSL20}.
Other researchers perturb the original image or extracted features through steganography and adversarial noise by generating visually obfuscated but machine-recognizable images \cite{DBLP:conf/mm/YuanLPLL022}, or by creating imperceptible visual perturbations to mislead attackers during reconstruction \cite{DBLP:conf/cvpr/00010JZHWSYLR23} or unauthorized recognition \cite{DBLP:journals/pami/LyuJHPLD23}.
These methods provide privacy to data before release. 
Under unlearning, some of the sensitive data will have already been used for training, too late for these techniques to be applied. Our study of machine unlearning focuses on privacy protection post-release of private information.


\noindent\textbf{Image Generation from Scene Graphs.}
\label{subsec:rw-sg}
Scene graphs provide structured representation of visual or textual scenes, capturing objects, attributes, and their relationships \cite{DBLP:journals/pami/ChangR00C023}.
Scene graph studies can be divided into two categories, scene graph generation and applications of scene graph \cite{DBLP:journals/pami/ChangR00C023}. Both areas have advanced for computer vision and natural language processing applications. 
Image generation from scene graphs methods often follow a layout-based image generation \cite{johnsonGF18, DBLP:conf/iclr/MittalAAMM19, DBLP:journals/corr/abs-1901-03762, DBLP:conf/cvpr/ZhaoMYS19, DBLP:conf/cvpr/DhamoFLNHT020}, 
in which two key sub-processes are scene layout generation~\cite{DBLP:conf/iccvw/SchroederTT19, DBLP:conf/eccv/HerzigBXCDG20, DBLP:conf/cvpr/ChaiZY23} and image generation from layouts~\cite{DBLP:conf/iccv/ChenK17, DBLP:conf/cvpr/Park0WZ19, DBLP:conf/cvpr/ZhengZLQSL23}. 
Among these studies, Chang \textit{et al.} \cite{DBLP:conf/cvpr/DhamoFLNHT020} provide a standardized framework integrating these core techniques, from which we shall construct the image generator model backbone.

\begin{figure*}
    \centering
     \includegraphics[width=\textwidth]{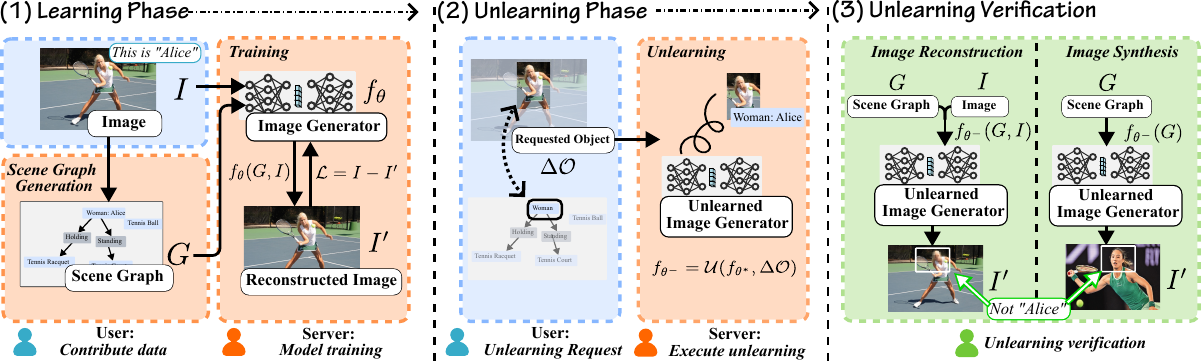}
     \vspace{-3mm}
    \caption{Scene graph-based object unlearning framework. Scene graphs can help both servers and users manage unlearning requests effectively by providing a structured way to understand the relationships between objects in an image. Scene graphs make it easier to identify and remove the requested object, such as a girl in an image. Moreover, this ensures that servers interpret and handle requests accurately, avoiding vague or incomplete unlearning actions. In this way, scene graphs act as a bridge, translating user intentions into actionable and transparent operations for the server. }
    \label{fig:scheme}
\end{figure*}

\section{Preliminaries}
In this section, we introduce the preliminary knowledge, notations, and settings in this study. 
A summary notation table is provided in Table~\ref{tab:notations}.


\begin{table*}[ht]
\centering
\caption{Notations.}
\label{tab:notations}
\resizebox{0.9\linewidth}{!}{%
\begin{tabular}{cl|cl|cl}
\toprule
\textbf{Notation} & \textbf{Explanation} & \textbf{Notation} & \textbf{Explanation} & \textbf{Notation} & \textbf{Explanation} \\ \midrule
$I$ & Image & $G \in \mathcal{G}$ & Scene graph, $G = (\mathcal{O}, \mathcal{E}) $ & $\mathcal{D}$ & Training data \\
$I'$ & Generated image & $\mathbf{o} \in \mathcal{O}$ & Object, $\mathbf{o} = (\mathbf{c}, \mathbf{a})$ & $\Delta \mathcal{D}$ & Removed (unlearned) data \\
$\mathbf{x}$ & Input to model generator & $\mathbf{c} \in \mathcal{C}$ & Category of object & $\mathcal{D} \backslash \Delta\mathcal{D}$  & Remaining data \\
$\mathbf{q}$ & Query for image retrieval & $\mathbf{a} \in \mathcal{A}$ & Attribute of object & $f_{\theta^*}$ & Original model\\
$S$ & Similarity function & $\mathbf{r}_{ij} \in \mathcal{R}$ & Relationship between objects $\mathbf{o}_i$ and $\mathbf{o}_j$ & $f_{\theta^-}$ & Unlearned model\\
& & $\mathcal{E}$ & Edge set, $\mathcal{E}=\{(\mathbf{o}_{i}, \mathbf{r}_{ij}, \mathbf{o}_{j})\}$ & $\mathcal{U}$ & Unlearning algorithm
 \\ \bottomrule
\end{tabular}
}
\end{table*}





\noindent\textbf{Image Generation from Scene Graphs.}
In this work, we assume that the target model is trained for image generation as the learning task. 
Generally, generation refers to the process of creating an image $I$ from a given input $\mathbf{x}$. The input $\mathbf{x}$ can be set of images, text descriptions, latent variables, and/or prompts. The generation process can be modeled as a function $f_{\theta} : \mathbf{x} \rightarrow I$, where $\theta$ are the parameters of a trained generation model. 
Image generation models effectively learn the mapping from the input space to the image space.
This process is often accomplished using generative models such as Generative Adversarial Networks (GAN), Variational Autoencoders (VAE), and Diffusion Models (DM).

Specifically, we consider \textit{image generation from scene graphs}, whereby the model holder performs both training and subsequently unlearning. 
Our technique applies scene graphs, and as such we do not explore scene graph generation techniques. 
We assume that every image for training or otherwise, 
will invoke an established algorithm to generate a scene graph. 
An overview of scene graph generation techniques are provided in Section~\ref{subsec:rw-sg}. 
We include two distinct image generation from scene graph tasks during our evaluation, \textit{image reconstruction} and \textit{image synthesis}. These tasks will be detailed in Section~\ref{sec:unl_veri}. 


\begin{figure}
    \centering
    \includegraphics[width=0.8\linewidth]{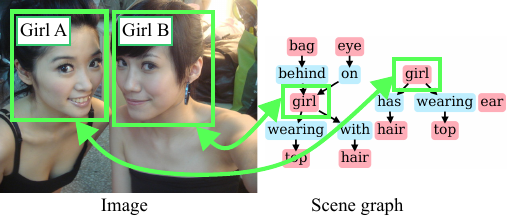}
    \caption{Illustration of the unique identity of objects in the scene graph. In the image and the corresponding scene graph, even though two objects may belong to the same category, such as ‘girl,’ they are represented as distinct objects.}
    \label{fig:obj_intro}
\end{figure}

\emph{Scene Graphs.} A scene graph is a data structure used to represent the contents of a scene by encoding objects, their attributes, and the relationships between the objects.
We use visual scene graphs (VSG) \cite{DBLP:journals/pami/ChangR00C023} to model images for object unlearning. 
Formally, given an image $I$, we have a corresponding scene graph
$G$.
Each scene graph $G$ is defined as a tuple $G = (\mathcal{O}, \mathcal{E})$, where $\mathcal{O} = \{\mathbf{o}_{1}, \dots, \mathbf{o}_{m}\}$ is a set of objects (nodes), $\mathcal{R}$ a set of relationship types,  and $\mathcal{E} \subseteq \mathcal{O} \times \mathcal{R} \times \mathcal{O}$ is a set of edges of the form $(\mathbf{o}_{i}, \mathbf{r}_{ij}, \mathbf{o}_{j})$ where $\mathbf{o}_{i}, \mathbf{o}_{j} \in \mathcal{O}$. Each object $\mathbf{o}_{i}$ can be expressed as $\mathbf{o}_{i} = (\mathbf{c}_{i}, \mathbf{a}_{i})$, where $\mathbf{c}_{i} \in \mathcal{C}$ is the category of the object, typically determined by $\mathbf{a}_{i} \subseteq \mathcal{A}$ which represents the attributes of the object. 
From the perspective of object, each object is \textit{unique} in identity; that is, given two objects $\mathbf{o}_{i}$ and $\mathbf{o}_{j}$, even if $\mathbf{c}_i = \mathbf{c}_j$ and $\mathbf{a}_i = \mathbf{a}_j$, it still holds that $\mathbf{o}_i \neq \mathbf{o}_j$ (consider a pair of twins). We use the notation $o \in I$ to say that the object $o$ is contained in the image $I$. The notation $\text{cat}(o)$ means the category of object $o$. Define the sets $O = \{o : o \in I \text{ for some } I \in \mathcal{D} \}$ and $\mathcal{C} = \{c : c = \text{cat}(o) \text{ for some } o \in O \}$, as the set of objects and categories in the training dataset, respectively.

Figure~\ref{fig:obj_intro} is an illustration of a scene graph, observe how the graph describes the hierarchical relationship between different objects in the image, not only is the ``bag'' object present in the image, the graph capture's it's position behind the ``girl'' on the right. As each graph node has it's own features, the two ``girl'' nodes are similar, but distinctly unique.

        

        
        

    

\noindent\textbf{Machine Unlearning.} \label{sec:def_unl}
\noindent\emph{The Setting.} We consider a MLaaS provider with complete control over their model, including its training data and white-box information. This trained model is accessible via an API for end users.
When compelled to remove a specific sensitive items from the model, and associated training data, the requester specifies the data/objects to be removed. For example, data contributors to a social media platform are automatically opted in for machine learning training of a friend photo tagging system, however a privacy conscious user can request the platform to unlearn their specific data from trained models, allowing the affected user to opt out.


\noindent\emph{Definitions.} 
Machine unlearning refers to the process of removing the influence of specific data points from a trained machine learning model. Let $\mathcal{D}$ be the full training dataset. We use the notation $f_{\theta^*}$ to denote the \textit{original model} trained on $\mathcal{D}$. Let $\Delta \mathcal{D}$ be the data to be removed from $\mathcal{D}$, and $\mathcal{D} \backslash \Delta\mathcal{D}$ be the remaining data after removing $\Delta\mathcal{D}$.
$\Delta\mathcal{D}$ is usually reflected in the unlearning request $\mathbf{q}_\mathrm{unl}$ of the user, which will be sent to the service provider for an unlearning execution $f_{\theta^-} = \mathcal{U}(f_{\theta^*}, \Delta\mathcal{D})$ where $\mathcal{U}$ is an unlearning algorithm and $f_{\theta^-}$ denotes the unlearned model.
Based on the nature of unlearning requests, there are four types of prevailing machine unlearning techniques: (1) sample unlearning,
(2) feature unlearning, (3) class unlearning, and (4) client unlearning. The unlearning techniques differences are summarized in Table~\ref{tab:unl_type}, together with our proposed object unlearning technique.

It is important to note that these five types of unlearning techniques have varying scopes  depending on the scenario. For instance, class unlearning is more likely to occur in discriminative tasks rather than generative tasks. 
Additionally, client unlearning is more relevant in distributed systems, such as federated learning. In this work, we focus primarily on generative tasks; therefore, our investigation centers on sample unlearning and feature unlearning. Consequently, we compare our proposed object unlearning method with these approaches. Below, we provide definitions of sample unlearning and feature unlearning in the context of generative tasks \cite{DBLP:conf/iccv/GandikotaMFB23, DBLP:conf/aaai/MoonC024}. 
These unlearning techniques are defined in terms of item sets from the unlearning request issued to the service provider,
$\mathbf{q}_\text{unl} \subseteq O$,
i.e., the subset of objects to be removed from the training dataset $\mathcal{D}$. 

\begin{definition}[Sample Unlearning]
Sample unlearning is defined as the removal of a specific data sample or a set of data samples from a generative model. Given a set of requested objects $\mathbf{q}_\text{unl}$, sample unlearning seeks to unlearn
\[
\Delta \mathcal{D} = \{ I \in \mathcal{D} : o \in I, \text{ for any } \mathbf{o} \in \mathbf{q}_\text{unl} \},
\]
i.e., all images containing one or more objects from the set $\Delta O$. 
For instance, if a specific image or group of images is unlearned from a generative model, the model would attempt to generate outputs that are not influenced by the characteristics of those unlearned images.
\end{definition}

\begin{definition}[Feature Unlearning]
Feature unlearning is defined as the unlearning technique where a specific feature is removed from a generative model. Given the set of objects $\mathbf{q}_\text{unl}$ as defined in the unlearning request, the goal of feature unlearning is to unlearn:
\[
\Delta \mathcal{C} = \{ c \in \mathcal{C} :  c = cat(o), \text{ for any } \mathbf{o} \in \mathbf{q}_\text{unl} \}.
\]
For example, after unlearning the ``boy'' feature from a generative model, the model would never generative images including visual feature recognized as a boy.
\end{definition}

We remark that we define feature unlearning from the perspective of the objects' features (e.g., man, tree, and sky) as investigated in \cite{guo2022efficient, DBLP:conf/iccv/GandikotaMFB23, DBLP:conf/aaai/MoonC024}. Another type of feature unlearning is related to overall image style (e.g., \textit{Van Gogh} style) \cite{DBLP:journals/corr/abs-2402-11846}, which is orthogonal to this study and will therefore not be discussed.


\section{Object Unlearning}
We now formulate object unlearning and detail an overall framework for scene graph-based object unlearning.


\noindent\textbf{Definitions.} \label{sec:obj_unl}
%
%
%
Recall the real-world setting of image-based machine learning tasks, specific information from images may have their removal requested, without requirements for the whole image to be removed. In this scenario, the limitations of sample and feature unlearning approaches are clear, much more is removed than what is required. Motivating our proposal for \emph{object unlearning}. 

\begin{definition}[Object Unlearning]

Object unlearning is defined as the unlearning technique where a specific object is removed from a generative model. Given the set of requested objects $\mathbf{q}_\text{unl}$ for unlearning, we unlearn:
\[
\Delta \mathcal{O} = \{ o \in O :  o \in \mathbf{q}_\text{unl} \}.
\]
For instance, given a specific requested unlearning object, e.g. a boy named ``Tom'', only this specific object is selectively removed or ``unlearned'' from the generative model. 
After unlearning, the model should not generate images containing the visual object identified as ``Tom''. 
\end{definition}

Object unlearning allows for the selective removal of a distinct and identifiable object (such as a specific individual or item) from a generative model. This level of granularity ensures that the model can unlearn highly specific visual or conceptual entities while retaining other related features or objects from the training sample. In other words, the model does not remove other objects in the same image. 



\noindent\textbf{Unlearning Verification Metrics.} \label{sec:unl_veri}
To assess whether object unlearning is successful in its task, we construct several metrics. 
For generative models, successful unlearning is expected to achieve three quantitative objectives in the unlearning verification phase: \textit{Effectiveness}, \textit{Preservation}, and \textit{Generalizability} \cite{DBLP:journals/corr/abs-2407-20516}. 
Under the context of our target for object unlearning, as discussed above, we formulate these two objectives as follows: 
\begin{itemize}[leftmargin=*, itemsep=0pt, topsep=0pt]
    \item \textbf{Unlearning Effectiveness:}   
    The unlearned model ($f_{\theta^{-}}$) should not generate the removed object ($\Delta \mathcal{O}$) in its generation. That is, for any input $\mathbf{x} = (I, G)$, which includes an image $I$ and its scene graph $G$, we require that:
    \begin{equation}
        \text{For all } \mathbf{o} \in \Delta O, \mathbf{o} \notin f_{\theta^{-}}\left(\mathbf{x}\right)
        \label{eq:veri_acc}
    \end{equation}
    
    \item \textbf{Model Utility Preservation:} The unlearned model ($f_{\theta^{-}}$) should maintain its performance on the retained objects in its generation. That is, for any input $\mathbf{x} = (I, G)$, we require that
    \begin{equation}
    \text{For all } \mathbf{o} \in \mathbf{x} \text{ such that } \mathbf{o} \in O\backslash\Delta O, \mathbf{o} \in f_{\theta^{-}}\left(\mathbf{x}\right)
    \label{eq:veri_local}
    \end{equation}
\end{itemize}

The objectives above are generalized to capture that the model achieves both Effectiveness and Preservation in the tasks of (1) image reconstruction and (2) image synthesis. 
These two image generation tasks are important for verification in different application settings.
For image reconstruction, the focus is to provide an exact evaluation to ensure that the information related to the requested object has truly been removed, given that the input still includes strong associated visual features of the original object samples.
In contrast, the image synthesis task is more relevant to realistic scenarios in generative AI applications. Here, the goal is to prevent objects containing personally identifiable information (PII), such as faces, from appearing in generations created by other users. While other scenarios may be constructed, we discuss them as future work in our discussion.

\subsection{Scene Graph-Based Object Unlearning Framework}

Object unlearning presents two significant technical challenges. The first is accurately identifying a distinct and recognizable object from the unlearning request, especially when similar or semantically close objects are present in the image. Specifically, the question of how we can reliably pinpoint the unique object in question.
The second challenge involves the disentanglement of interconnected objects, ensuring that the removal of one element does not unintentionally diminish the model's understanding or introduce inconsistencies in the remaining structure. 
To address this, we propose a scene graph-based object unlearning framework.

\emph{Framework Overview:}
The proposed scene graph-based object unlearning framework contains integrations within both learning and unlearning phases, presented in Figure~\ref{fig:scheme}.

\begin{figure}[th]
    \centering
    \includegraphics[width=0.9\linewidth]{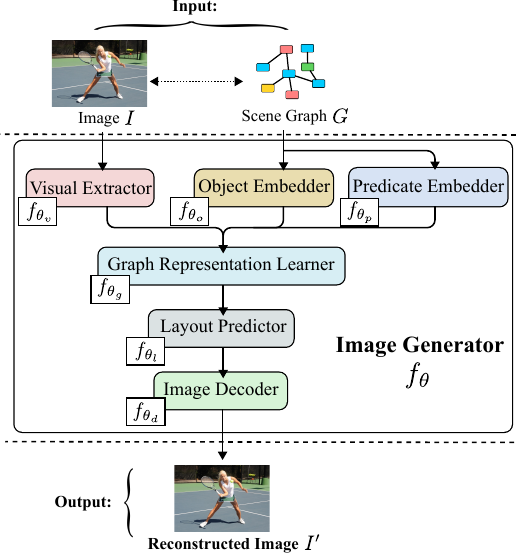}
    \caption{Schematic of scene-graph-to-image (SG2I) generator. Readers can refer to \cite{DBLP:conf/cvpr/DhamoFLNHT020} for a detailed view of the architecture.}
    \label{fig:g2i}
\end{figure}

\emph{Learning Phase:}
During the learning phase, the MLaaS provider and data provider (user) collaboratively train a image generator\footnote{In this paper, ``image generator'' and ``generative image model'' are used interchangeably.}
During learning phase, a scene-graph-to-image (SG2I) generator~\cite{johnsonGF18} as the image generator $f_{\theta^*}$ is trained. 

A architecture of this family of generators is illustrated in Figure~\ref{fig:g2i}.
This SG2I generator takes as input both images and their corresponding scene graphs, i.e., $\mathbf{x}=(I, G)$, for model training. 
As stated before in our assumptions, the service provider will generate corresponding scene graphs when new data samples are contributed. We assume said scene graphs exists.
For example, platforms like Google Cloud can automate processes like object detection, scene understanding, and tagging without user intervention\footnote{https://cloud.google.com/blog/products/ai-machine-learning/label-your-photos-automagically-with-vision-api/}. 
With both images and scene graphs available, the server will train the SG2I generators to learn how to map scene graphs to images. 
As SG2I stores the visual feature information of objects, it will later become the focus for object unlearning.

After training, we obtain a trained model $f_{\theta^*}$ (the original model on which unlearning is to be applied), formulated for a given learning task ($\mathcal{L}_{\mathcal{D}}$), i.e., image reconstruction, we have:
\begin{equation}
\theta^*=\underset{\theta}{\arg \min } \mathcal{L}_{\mathcal{D}}, \quad \mathcal{L}_{\mathcal{D}}=\sum_{I_i \in \mathcal{D}} l\left(f_{\theta}\left(G_i, I_i\right), I_i\right) \label{eq:learn}
\end{equation}
where $l$ is the loss function defined on the reconstruction of each image $f_{\theta}\left(G_i, I_i\right)$ and the ground truth $I_i$.

\emph{Unlearning Phase:}
In the event an unlearning request $\mathbf{q}_\mathrm{unl} = { \mathbf{o} \in \Delta \mathcal{O} }$ is made to the MLaaS provider to unlearn a specific object. The MLaaS provider will execute unlearning algorithm $\mathcal{U}$ to produce an unlearned model:
\begin{equation}
    f_{\theta^-} = \mathcal{U}(f_{\theta^*}, \Delta\mathcal{O}).
\end{equation}


\subsection{Object Unlearning Approaches} \label{sec:if_alg}


The proposed scene graph-based framework precisely identifies the object of interest within complex visual data.
Specifically, when constructing a scene graph, each object is assigned a unique \textit{bounding box} specifying its positional information within the image. This is beneficial by allowing the retrieval of precise regions of interest (ROI) of the object during the unlearning process. 
Once these objects are clearly defined, we can effectively apply a targeted unlearning algorithm to remove them.

In this section, we redeploy three efficient approximate unlearning methods to serve as alternatives to the computationally intensive approach of retraining the model sans the unlearning object. 
Methods 1 and 2 employ fine-tuning techniques, while Method 3 leverages influence functions for model redaction, also known as model editing. In subsequent experiments, we evaluate the effectiveness and efficiency of these approaches in achieving targeted object unlearning.

\noindent\textbf{Methodology 1: Negative Guidance-Based Fine-Tuning.} 
We first propose a \textit{negative guidance-based fine-tuning} method.
In each training iteration, for the specific target object, we first locate its bounding box in the scene graph to extract the corresponding region of interest (ROI) of the target object. 
Then, a reconstruction loss is computed between the corresponding ROI areas of the generated image and the target image. 
To achieve unlearning, we negate this loss and add it to the total loss as a negative guidance term.
The negative guidance loss is defined as follows:
\begin{equation}
\mathcal{L}_{\text{ng}} = - \lambda \cdot \sum_{I_i \in \mathcal{D}} l\left(I'_{i,\mathbf{o}}, I_{i,\mathbf{o}}\right)
\end{equation}
Here, $I'_{i,\mathbf{o}}$ and $I_{i,\mathbf{o}}$ denote the ROI of the generated and original images, respectively; $\lambda$ is a weighting factor that balances the influence of negative guidance with the generative objective; and $l$ represents the reconstruction loss, which calculates the pixel-wise difference.

This loss function leads the generator to gradually remove the feature representation of the object. Finally, we add the negative guidance loss to the total generator loss, which guides the generator and updates the parameters to unlearn the target object:
\begin{equation}
\mathcal{L}_{\text{total}} = \mathcal{L}_{\text{gen}} + \mathcal{L}_{\text{ng}}
\end{equation}
where $\mathcal{L}_{\text{gen}}$ represents the generator’s original loss function. This process will weaken the generator's memory of the target object gradually, thereby realizing object unlearning.

\noindent\textbf{Methodology 2: Mask-Based Fine-Tuning.}
A \textit{mask-based fine-tuning} process involves two main steps: (1) mask the ROI associated with the requested object $\mathbf{o}$, and (2) fine-tune the model with this masked input.

Let $M_\mathbf{o}$ be a mask that covers the region associated with the object $\mathbf{o}$ of the scene graph in the image $I$. Using the bounding box information provided by the scene graph, this mask can be easily localized and constructed. Further, we can obtain a modified input $\tilde{I} = I \circ M_\mathbf{o}$, where $\circ$ here denotes element-wise masking, ensuring that only the region associated with $\mathbf{o}$ is influenced while preserving the remainder of the image.
Particularly, we introduce two types of masks $M_\mathbf{o}$ for ROI as follows (we use $x$ and $y$ to denote the pixel coordinates below):
\begin{itemize}
    \item \textit{Patch Masking}: Set pixel values in $M_\mathbf{o}$ to zero:\\
    $\tilde{I}_{x, y} = 0 \quad \forall \, x_{\text{left}}' \leq x \leq x_{\text{right}}', \, y_{\text{top}}' \leq y \leq y_{\text{bottom}}'$.
    \item \textit{Noise Masking}: Inject Gaussian or random noise $\mathcal{N}(0, \sigma^2)$ to the ROI covered by: \\$M_\mathbf{o}$, $\tilde{I}_{x, y} = I_{x, y} + \eta, \quad \eta \sim \mathcal{N}(0, \sigma^2)$.
\end{itemize}
To unlearn the object $\mathbf{o}$, the model $f_{\theta}$ is fine-tuned with the modified $\tilde{I}$. This process updates the model parameters from $\theta$ to $\theta^-$ by minimizing a loss function $L$ that measures the model's output consistency with the original unmasked regions in $I$:
\begin{equation}
\theta^- = \arg \min_{\theta} \sum_{I_i \in \mathcal{D}} l\left(f_{\theta}\left(G_i, \tilde{I}_i\right), \tilde{I}_i\right)
\end{equation}
where $\tilde{I}_i$ represents each instance of a masked input. 
This design offers a straightforward solution to guide the model in retaining unmasked features.
After fine-tuning, the adjusted model $f_{\theta^-}$ should avoid generating the object $\mathbf{o}$ in future outputs while maintaining other objects in the image.

\noindent\textbf{Methodology 3: Influence Function-based Partial Model Redaction.} \label{sec:method_ifpe}
Influence functions permit the approximation of the unlearning process, thereby achieving efficient unlearning. 
Specifically, our scene graph-based object unlearning can be reformulated as a graph unlearning problem~\cite{said2023survey}. As each object within the image directly corresponds to a node within the scene graph, we first formulate the object unlearning task as a node-level graph unlearning problem.

To solve this problem, we draw upon the off-the-shelf work of \cite{wu2023gif}, which explores the use of influence functions for node-level graph unlearning. In \cite{wu2023gif}, the authors provide a proven closed-form expression for the model parameter change, $\Delta \theta = \theta^{-} - \theta^{*}$, which is applicable to our scenario. The expression is given by:
\begin{equation}
\Delta \theta \approx H_{\theta^{*}}^{-1} \nabla_{\theta^{*}} \mathcal{L}_{\Delta \mathcal{O}},
\end{equation}
where $H_{\theta^{*}}$ is the Hessian matrix of the learning loss $\mathcal{L}_{\mathcal{D}}$ concerning $\theta^{*}$.

To properly account for the unlearning of specific objects $o \in \Delta \mathcal{O}$, we leverage knowledge that a scene graph object corresponds to a node with its own attributes (e.g. label, identity, or location). Consequently, object unlearning can be framed as node feature unlearning within the broader graph unlearning landscape. 
Finally, the unlearned model can be estimated by model redaction:
$\theta^{-} = \theta^{*} + \lambda \Delta \theta \label{eq:redact}$,
where $\lambda$ is a scalar multiplier that adjusts the magnitude of the parameter change. 
We will give details of this method in Appendix~\ref{app:method_mpr}

\section{Experimental Setting}
In this section, we first introduce the dataset, model, and evaluation metrics employed in our experiments. Followed by a presentation of the learning and unlearning settings.


\noindent\textbf{Dataset.}
The Visual Genome dataset \cite{vg} is a large-scale resource designed to advance research in image understanding, particularly in tasks like object detection, scene recognition, and relationship modeling. It contains $108,077$ images annotated with approximately $21.3$ million object instances, $10.8$ million attributes, and $1.5$ million relationships. Additionally, it provides $5.4$ million region descriptions and $1.7$ million question-answer pairs. This dataset is instrumental in tasks such as scene graph generation, visual question answering, and image captioning, making it a critical benchmark \cite{DBLP:journals/ijon/LiZZJDHSZSB24}.

\noindent\textbf{Pre-processing.}
To process the dataset, we develop a pipeline for both our training and unlearning processes.

For object processing, we construct vocabularies encompassing objects, attributes, and relationships. We standardize the naming conventions for objects and relationships using alias mappings to ensure consistency. Once the vocabularies are established, we filter the object annotations to retain only those that met specific size criterion and are included in the constructed vocabulary. Additionally, objects and attributes that appear frequently (above a predetermined threshold) were also incorporated into the vocabulary.

For image processing, consistency was ensured by removing images with dimensions below a specified minimum size, particularly those with extremely small objects.
With realistic privacy-preserving scenarios in mind, we specifically select all samples containing salient personally identifiable information (PII).
This was achieved by identifying and including all samples labeled with any of the following nine human-related object labels: [``$\mathrm{man}$'', ``$\mathrm{woman}$'', ``$\mathrm{boy}$'', ``$\mathrm{girl}$'', ``$\mathrm{child}$'', ``$\mathrm{person}$'', ``$\mathrm{kid}$'', ``$\mathrm{people}$'',  ``$\mathrm{face}$''].

We then encoded the objects, attributes, and relationships into a scene graph-based representation for each image. 
To ensure uniformity across all images, the data were padded to maintain a consistent structure, with each image containing a specified number of objects ($|\mathcal{O}| = 10$) and relationships.

\noindent\textbf{Model.}
In our experiment, we employ SIMSG \cite{DBLP:conf/cvpr/DhamoFLNHT020} as the SG2I generator. SIMSG provides a general framework that has been widely adopted, as illustrated in Figure~\ref{fig:g2i}. This framework integrates VGG \cite{DBLP:journals/corr/SimonyanZ14a} as the visual feature extractor, a graph convolution-based heterogeneous GNN as GRL, and SPADE \cite{DBLP:conf/cvpr/Park0WZ19} as the image decoder. Due to computing resource limitations, the SG2I generator processes images at a resolution of $64 \times 64$ pixels for both input and output.

For training the SG2I model, we first pretrain the entire model on the whole training dataset. Following this, we fine-tune the model on samples containing the selected human-related object labels for $2000$ epochs to ensure the original model possesses sufficient generation capability.

\noindent\textbf{Metrics.}
In the verification of the unlearning framework, we use the metrics of MAE, SSIM, and LPIPS \cite{LPIPS} to measure the quality of unlearning, metrics common among related works \cite{DBLP:journals/corr/abs-2402-11846, DBLP:conf/aaai/MoonC024}.
Generally speaking, smaller MAE and LPIPS, or higher values of SSIM indicate better recovery of the generated images when compared against the ground truth. 
Further, as objects have different sizes between samples, we apply normalization to the metrics as needed. 
For SSIM and LPIPS, we resize each object to the same dimension for calculating the scores.

These four basic metrics address the objectives of the object unlearning verification discussed in Section~\ref{sec:unl_veri}, however, as there are multiple occurring objects including those not subject to the unlearning request, we can further develop three dimensions of metrics for object unlearning:
\begin{itemize}[leftmargin=*, itemsep=0pt, topsep=0pt]
    \item \emph{A1: Removal of the unlearned objects.} We will compare the difference between the unlearned object generated by the original model and the unlearned model, to evaluate \textit{unlearning effectiveness} as defined in Section~\ref{sec:unl_veri}. Greater differences of this metric, indicate better unlearning performance of the requested object.
    \item \emph{A2: Preservation on the retained Objects.} By comparing the differences between ``the retained objects of the sample'' as generated by the original model and the unlearned model, we can evaluate \textit{model utility preservation}, as defined in Section~\ref{sec:unl_veri}. Smaller differences of this metric, mean better unlearning focus on the requested objects.
    \item \emph{A3: Preservation on the objects with the same category of the unlearned objects in other samples.} We will also compare the differences between ``the objects with the same category of the unlearning objects in other samples'' as generated by the original model and the unlearned model, an alternative perspective to evaluate \textit{model utility preservation}. The smaller this metric, the better the unlearning focus on the specific sample.
\end{itemize}
We will present the metric in an abbreviated form in the evaluation section. For example, ``$\mathrm{A1\_SSIM}$'' represents the SSIM between the unlearned objects generated by the original model and the unlearned model. These metrics are summarized in Table~\ref{tab:metric}. 
It is important to note that these three dimensions should not be viewed individually, but rather in unison, to assess the tradeoffs of the unlearning process. 

\begin{table}[ht]
\small
\centering
\caption{Object unlearning metrics evaluated across three dimensions. A downward arrow ($\downarrow$) indicates that a lower metric value signifies better performance, and vice versa.}
\label{tab:metric}
\resizebox{0.8\linewidth}{!}{%
\begin{tabular}{cc}
\toprule
\textbf{Dimension} & \textbf{Metric} \\ \midrule
A1 & A1\_SSIM$\downarrow$, A1\_LPIPS$\uparrow$, A1\_MAE$\uparrow$  \\
A2 & A2\_SSIM$\uparrow$, A2\_LPIPS$\downarrow$, A2\_MAE$\downarrow$  \\
A3 & A3\_SSIM$\uparrow$, A3\_LPIPS$\downarrow$, A3\_MAE$\downarrow$ \\
\bottomrule
\end{tabular}
}
\end{table}


\noindent\textbf{Image Generation Training Settings.}
As described earlier, the image generator must be sufficiently capable to generate the original image before unlearning is applied. As such, we fine-tune the image generation technique on a smaller subset of the whole training set of Visual Genome dataset, that is the focus of unlearning within this experimentation set. 

\noindent\textbf{Unlearning Baselines.}
As discussed in Section~\ref{sec:obj_unl}, we shall evaluate the effectiveness of object unlearning by comparing the proposed framework against existing unlearning methods for sample unlearning. For this purpose, we introduce five baselines plus our devised four object unlearning-dedicated methods, they are:
\begin{itemize}[leftmargin=*]
    \item \textbf{Sample-FT}: Fine-tune the model by excluding the sample containing the requested object from the training dataset.
    \item \textbf{Sample-NG}: Fine-tune the model by applying negative guidance on the sample containing the requested object to reduce its influence.
    \item \textbf{Feat-IF}: Employ the influence function to remove features associated with the requested object.
    \item \textbf{Feat-NG}: Fine-tune the model with negative guidance applied to specific features associated with the requested object.
    \item \textbf{Feat-MK}: Fine-tune the model with a mask applied to features related to the requested object to obscure them.
    \item \textbf{Obj-IF}: Use the influence function to directly remove the requested object from the model’s representation.
    \item \textbf{Obj-NG}: Fine-tune the model by applying negative guidance directly on the requested object to minimize its influence.
    \item \textbf{Obj-MK-PA}: Fine-tune the model with a patch mask applied to the feature area associated with the requested object, obscuring it within the model’s internal representation.
    \item \textbf{Obj-MK-NS}: Fine-tune the model with a noise mask applied to the feature area containing the requested object to disrupt its learned features.
\end{itemize}


It is important to note that the sample and feature unlearning methods listed above differ from the sample and feature unlearning \textit{requests} discussed in Section~\ref{sec:def_unl}.
Furthermore, we implement negative guidance and influence influence function for all sample, feature, and object unlearning by generally following the idea we propose in Section~\ref{sec:if_alg}. 
For fine-tuning-based methods, the fine-tuning process is set to run for $200$ epochs.
There are some small adaption when implementing for different cases.


\begin{figure*}[t]
    \centering
     \includegraphics[width=0.85\textwidth]{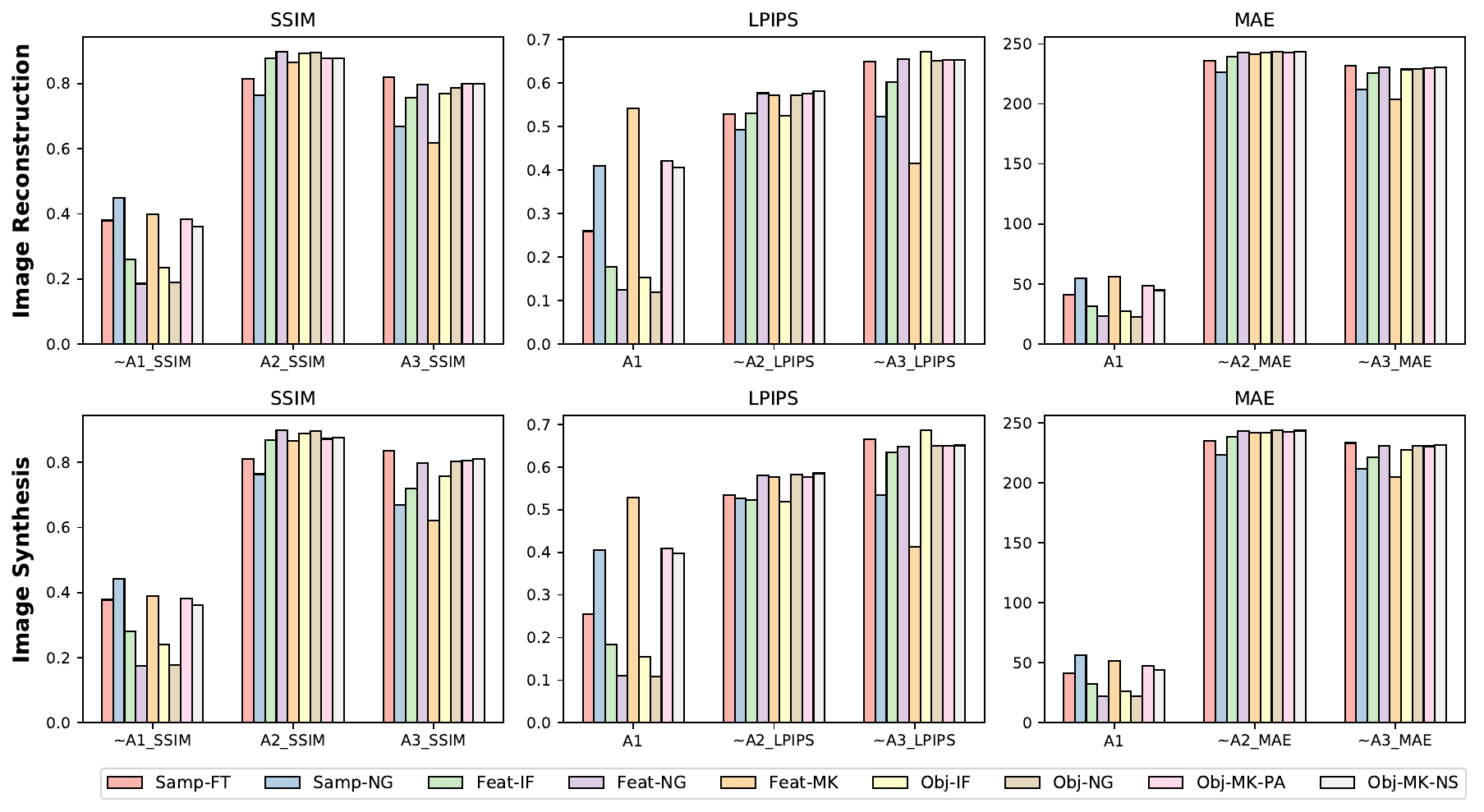}
    \caption{Results of unlearning verification through metrics regarding A1, A2, and A3. 
    To provide clarity for the reader, we have modify the distance metrics to follow a ``larger is better'' mantra. Specifically, we compute the complement values for A1\_SSIM, A2\_LPIPS, A3\_LPIPS, A2\_MAE, and A3\_MAE for presentation within the plot. 
    For SSIM and LPIPS, the complement transformation is 1 - value. For MAE, the complement transformation is 255 - value. The complement values are highlighted with a ``$\sim$'' prefix. For A3 metrics, since they involve multiple other samples, we calculate and report the average value among samples. 
    }
    \label{fig:exp_radar}
\end{figure*}

\begin{figure*}[t]
    \centering
    \begin{subfigure}[t]{0.49\textwidth}
        \centering
        \includegraphics[width=\textwidth]{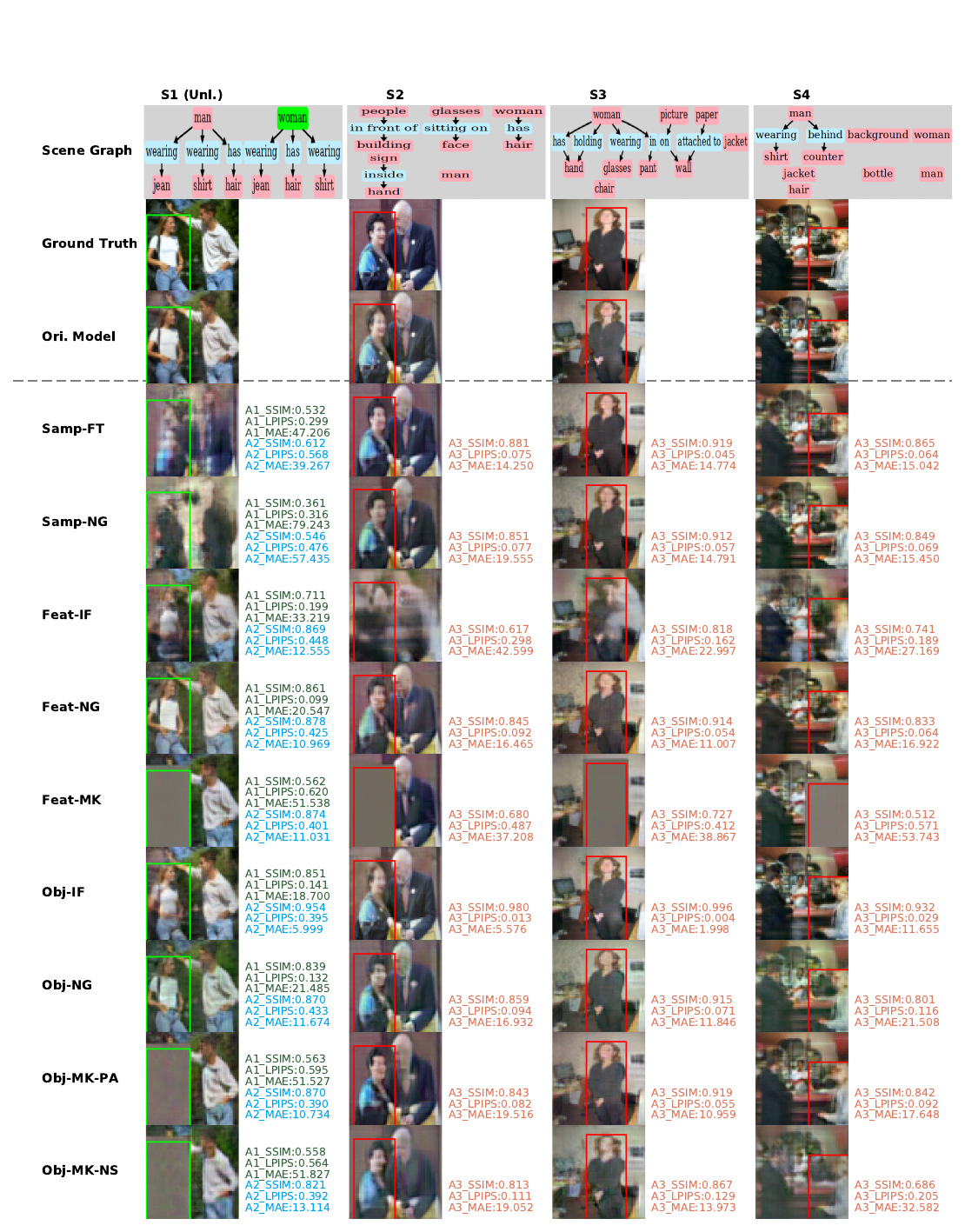}
        \caption{\textit{image reconstruction}.}
        \label{fig:exp_recon_1}
    \end{subfigure}
    \hfill
    \begin{subfigure}[t]{0.49\textwidth}
        \centering
        \includegraphics[width=\textwidth]{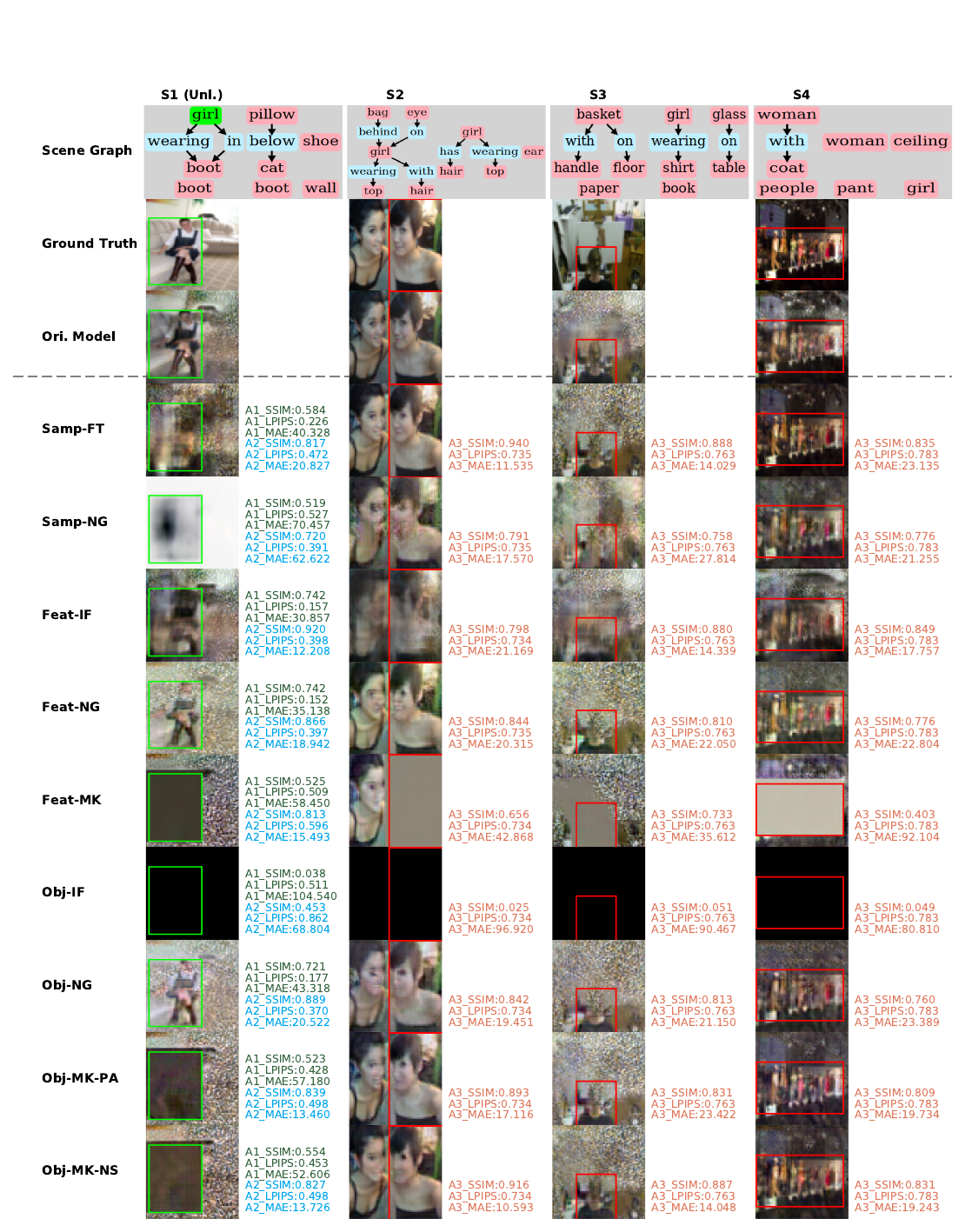}
        \caption{\textit{image synthesis}.}
        \label{fig:exp_syn_1}
\end{subfigure}
    \caption{Visualization of unlearning verification. In the images, the 'green boxes' localize the unlearned object, while the 'red boxes' indicate objects of the same category as the unlearned object but in different samples. In the scene graph visualization, the 'green node' represents the unlearned object. GT represents the ground truth.}
    \label{fig:exp_recon}
\end{figure*}


\begin{figure*}[t]
    \centering
    \begin{subfigure}[t]{0.49\textwidth}
        \centering
        \includegraphics[width=\textwidth]{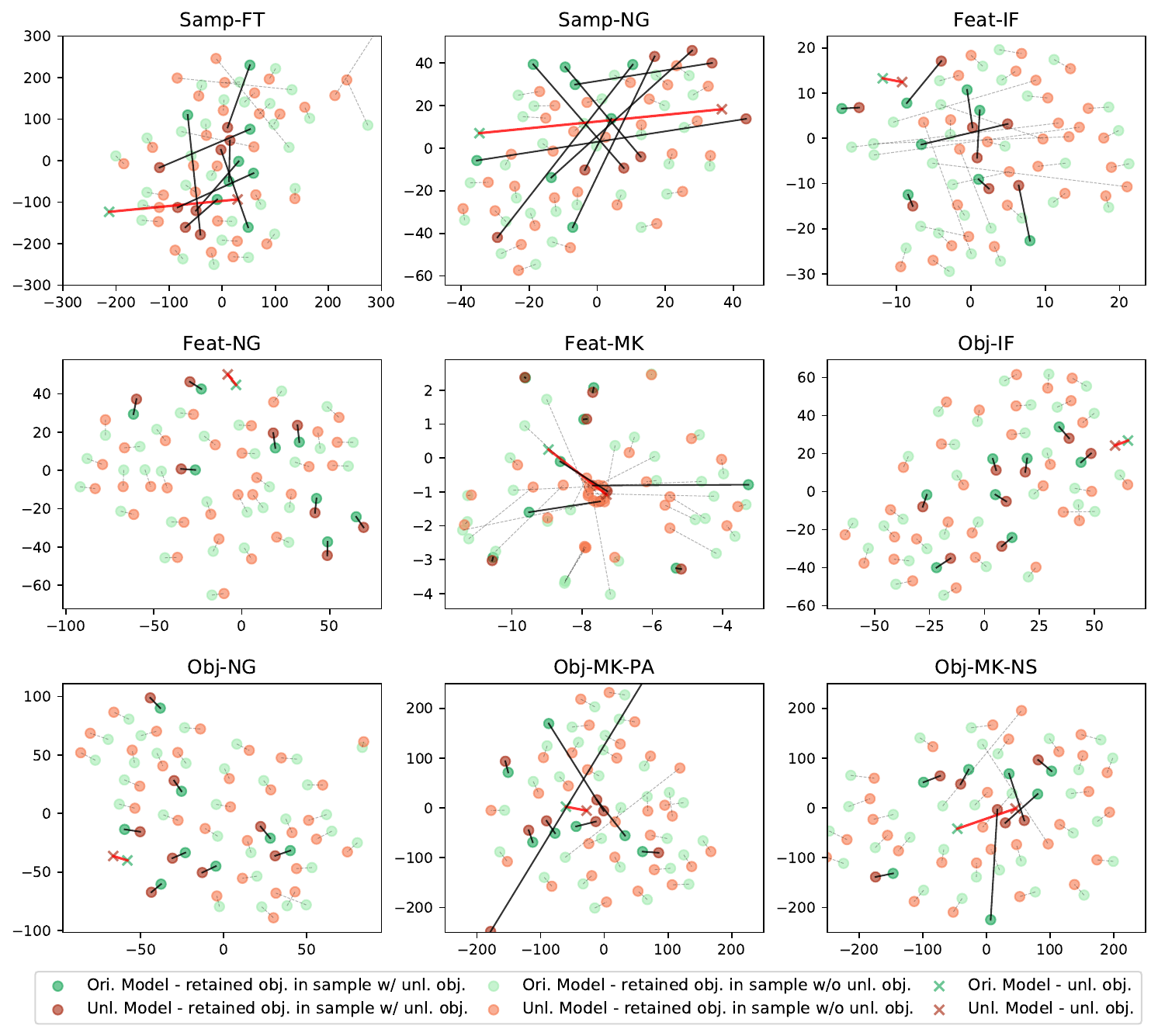}
        \caption{\textit{image reconstruction}.}
        \label{fig:exp_tsne_reconstruction}
    \end{subfigure}
    \hfill
    \begin{subfigure}[t]{0.49\textwidth}
        \centering
        \includegraphics[width=\textwidth]{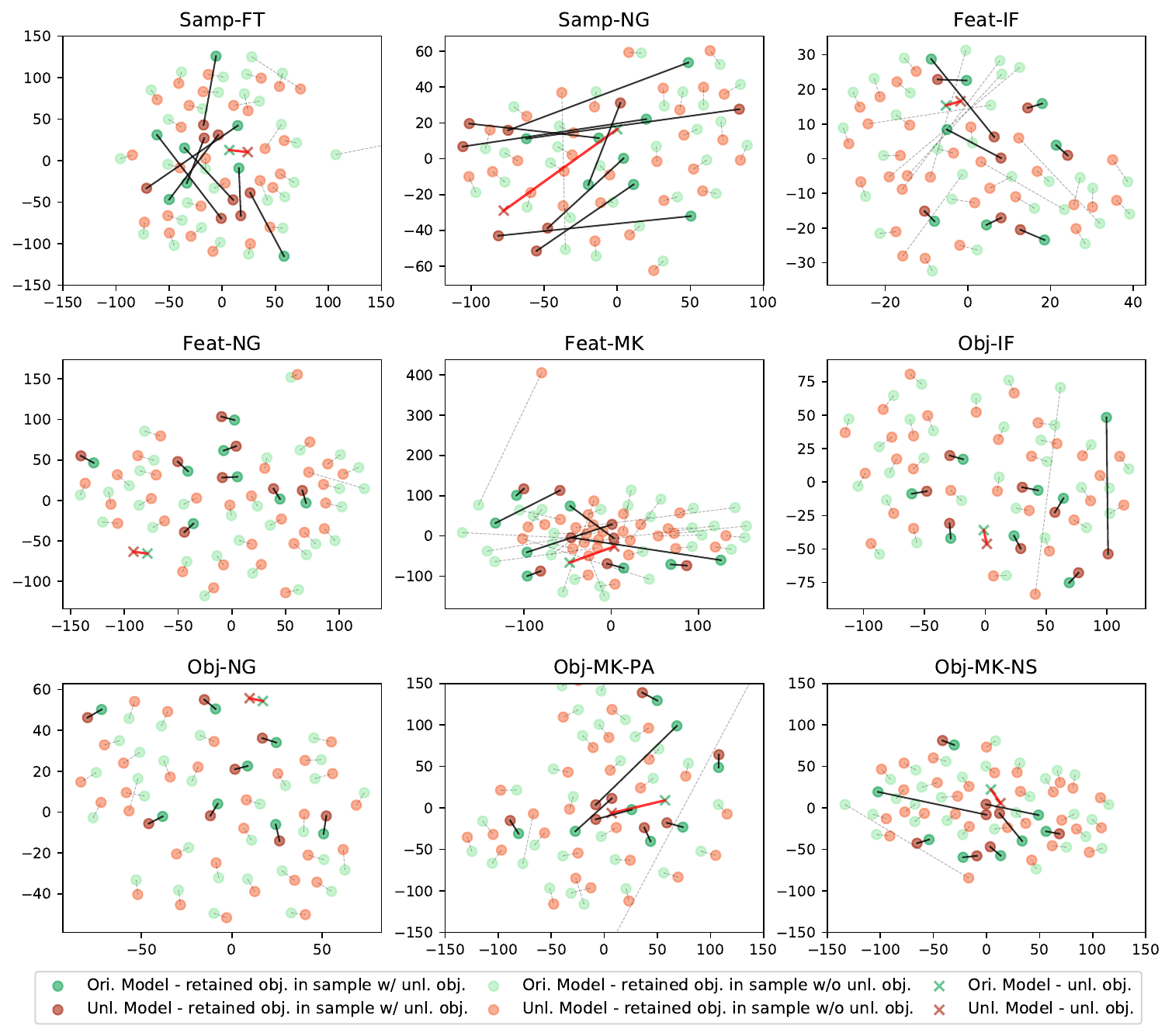}
        \caption{\textit{image synthesis}.}
        \label{fig:exp_tsne_synthesis}
\end{subfigure}
    \caption{The latent features of the object, as developed by the original model and the unlearned models through various unlearning methods in the image reconstruction task, are projected into a two-dimensional space using t-SNE. Our analysis confirms the effectiveness of unlearning methods in altering requested object representations while preserving others.}
    \label{fig:exp_tsne}
\end{figure*}



\section{Evaluation}
In this section we discuss findings first from unlearning in image reconstruction, followed by image synthesis.


\subsection{Unlearning in Image Reconstruction} \label{sec:exp_recon}
In evaluating the effectiveness of object unlearning, we employ image reconstruction as one key technique. Image reconstruction is selected for the inherent analogue it presents for our task of measuring object unlearning. Specifically, the input information used for reconstruction still exists within the object embeddings, making it a best case scenario for the model to recreate the original image. As such any impact of the unlearned object is due to the model no longer understanding this specific object. as such the poor restoration of the object, would indicates that the unlearning process has been highly effective. 
This method provides a stringent test of the model’s ability to selectively forget specific objects while retaining the integrity of other visual elements, thereby measuring unlearning success. We provide the metric results and a visualization in Figures~\ref{fig:exp_radar} (top half) and \ref{fig:exp_recon} (left half), respectively. 
It is worth mentioning that we will only demonstrate three additional samples of the training set in addition to the sample containing requested object due to the limit of space (i.e., S2, S3, S4 in Figure~\ref{fig:exp_recon}, hereinafter).
Some additional results of the visualization are provided in Figure~\ref{fig:exp_recon_syn_2} in Appendix~\ref{app:additional}.

\textbf{Observation 1:} Methods based on negative guidance and model redaction demonstrate poor performance. These approaches either fail to effectively forget the requested object or significantly compromise model utility.
In particular, the Obj-IF method, which we developed based on the existing influence function-based model redaction, demonstrated suboptimal performance. To further investigate the potential contributing factors, we conducted a detailed ablation study.
This suggests that in more complex generation tasks, negative guidance must be designed in a more innovative manner to effectively facilitate the forgetting process. 

\textbf{Observation 2:} All sample unlearning methods successfully eliminate the requested object; however, they also erase the remaining information from the original sample, resulting in a loss of model utility. This outcome aligns with our hypothesis that sample unlearning is limited to the sample level and struggles to achieve selective unlearning.

\textbf{Observation 3:} Among all the methods, the masking-based approach proves to be the most effective, i.e., Feat-MK, Obj-MK-PA, and Obj-MK-NS. As shown in the experimental results, the requested object is successfully removed while preserving other information in the image. This performance is notably superior to sample unlearning. Furthermore, unlike the feature unlearning method, objects with similar features in other images are retained. These results highlight the advantages of our approach, which enables selective unlearning with greater precision and efficiency.

\takeawy{Our method, Obj-MK-PA and Obj-MK-NS, demonstrates satisfactory results across A1, A2, and A3, achieving a satisfactory performance in terms of both unlearning effectiveness and model utility. This highlights the effectiveness of our approach in targeted unlearning, ensuring that the desired objects are forgotten while preserving the model’s utility on the remaining data.}

\subsection{Unlearning in Image Synthesis} \label{sec:exp_syn}
In evaluating the effectiveness of object unlearning, we also consider image synthesis, as task closely aligned with real-world scenarios encountered in MLaaS. In these environments, users often provide textual descriptions or prompt-based inputs, relying on the model to generate images entirely based on the learned information. Our evaluation method reflects this use case by utilizing scene graphs as the sole input for the SG2I process. By examining the model’s ability to generate images from scene graphs after specific objects have been unlearned, we can assess whether the model effectively forgets the targeted objects while still accurately reconstructing the remainder of the scene. 
The metric results and a visualization of this evaluation are present in Figures~\ref{fig:exp_radar} (bottom half)  and \ref{fig:exp_recon} (right half), respectively. Some additional results of the visualization are provided in Figure~\ref{fig:exp_recon_syn_2} in Appendix~\ref{app:additional}.

\textbf{Observation 1:} Overall, due to the stochastic nature of the synthesis process and the limited visual feature information available to the model, the unlearned object’s information is barely represented in any of the images generated by the unlearned model. This indicates that the unlearning process has been effective across all methods, as evidenced by the overall improvement in performance on A1.

\textbf{Observation 2:} In the image synthesis task, it is evident that some methods based on negative guidance and influence functions experienced catastrophic unlearning. The performance of the unlearned model showed a significant decline as a result. The inherent randomness in the image synthesis process affects their ability to generate images on other samples, leading to greater distortion compared to the original model, even when visually recognizable objects are generated. This distortion can be observed in the generation results in Figure~\ref{fig:exp_recon}. 
In contrast, our proposed Obj-MK-PA and Obj-MK-NS are less susceptible to this effect, maintaining the ability to generate visual features similar to those produced by the original model. The advantage of both methods is particularly pronounced in A2 and A3, as its focus on unlearning more specific information allows the SG2I model to recover the most accurate possible representation of the original image, leveraging its generative capabilities.

\vspace{1mm}
\takeawy{The generalizability of SG2I model in image synthesis introduces randomness that challenges traditional unlearning methods on retaining original visual information, leading to more distortion. In contrast, our proposed Obj-MK-PA and Obj-MK-NS show a more robust performance on maintaining the quality of the remaining content.}

\subsection{Unlearning Analysis in Latent Space}

We conducted a detailed comparison of the latent features of objects generated by the original model and the unlearned model. The results are visualized in Figure~\ref{fig:exp_tsne} for both image reconstruction and synthesis tasks. The analysis aligns well with the observed unlearning effects as discussed in Section~\ref{sec:exp_recon} and \ref{sec:exp_syn}. We have three major observations.

\textbf{Observation 1:} 
The methods Obj-MK-PA and Obj-MK-NS, which exhibited the best performance in unlearning, showed significantly larger latent space feature distances for the requested objects. In contrast, the distances for other objects in the same samples containing the requested object and for objects in samples without the requested object were relatively small. This confirms that these successful methods achieve effective unlearning by modifying the latent feature space of the requested object while maintaining the features of other objects.

\textbf{Observation 2:} 
Even among successful methods, we observed cases where the latent space feature distances for retained objects between the original and unlearned models were unexpectedly large. This is likely due to the inherent complexity of generative models, which can introduce random fluctuations in the latent space. These variations highlight the stochastic nature of generative processes and suggest that some unintended noise may affect the representation of retained objects.

\textbf{Observation 3:} 
Interestingly, we found little difference in latent space results between the image reconstruction and image synthesis tasks. This suggests a shared latent feature behavior across these tasks, despite their differences in objectives and output. However, further analysis is needed to better understand this consistency and its implications for unlearning in generative models.

\vspace{1mm}
\takeawy{Our analysis of latent space features shows clear evidence that the unlearning methods work effectively, especially in changing how requested objects are represented while keeping others intact. However, we still need to explore more to understand why there are differences in performance between image reconstruction and image synthesis tasks.}

\begin{figure*}[t]
    \centering
    \begin{subfigure}[t]{1\textwidth}
        \centering
        \includegraphics[width=.9\textwidth]{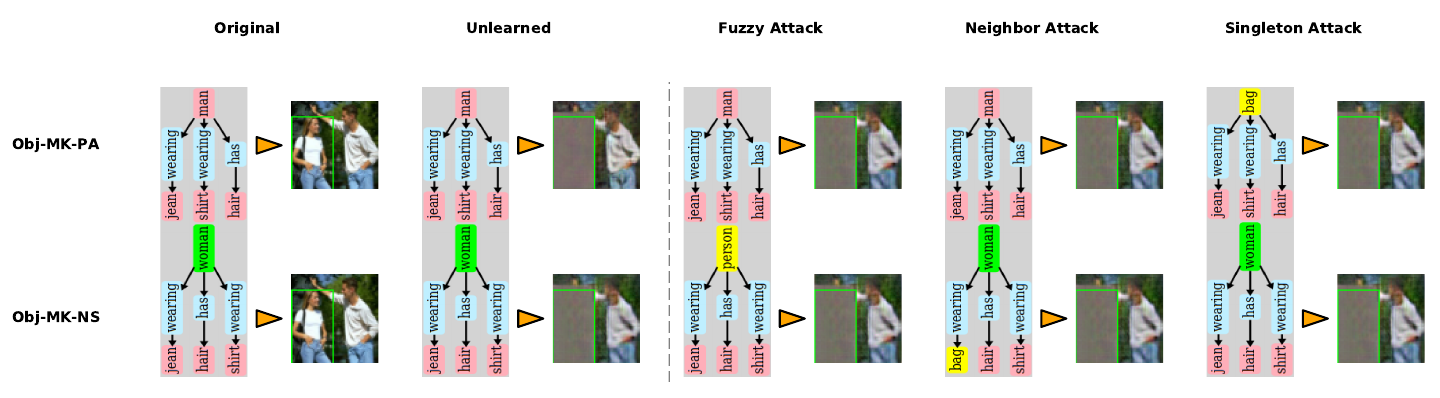}
        \caption{\textit{image reconstruction}.}
        \label{fig:exp_recon_leak}
    \end{subfigure}
    \hfill 
    \begin{subfigure}[t]{1\textwidth}
        \centering
        \includegraphics[width=.9\textwidth]{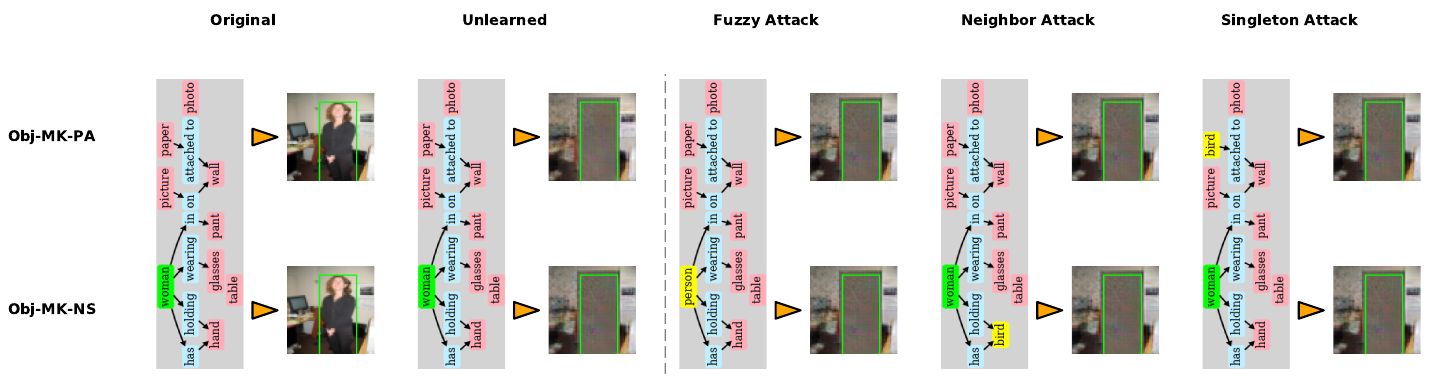}
        \caption{\textit{image synthesis}.}
        \label{fig:exp_syn_leak}
\end{subfigure}
    \caption{Indirect leakage tests. The modified object label in the scene graph is highlighted in 'yellow boxes'. It is clear that none of these three attacks can successfully make the model leak information about objects that have been unlearned.}
    \label{fig:indirect_leakage}
\end{figure*}



\begin{table}[ht]
    \caption{General comparison of average running time on implementing object unlearning. }
    \label{tab:run_time}
    \centering
   \resizebox{0.8\linewidth}{!}{%
        \begin{tabular}{lr}
        \toprule
        \textbf{Method} & \textbf{Average Running Time (s)} $\downarrow$ \\ \midrule
        Retrain & $13898.59 \pm $ $82.75$\\
        Fine-tune & $529.07 \pm 12.92$ \\
        Model Redaction & $3.25 \pm 0.72$ \\
        \bottomrule
        \end{tabular}
    }
\end{table}



\subsection{Indirect Leakage Test}
In this section we investigate if object unlearning sufficiently removes knowledge of the object from nearby scene graphs related to the original object. Effectively measuring the robustness of the unlearning process. To this end, we introduce three query variants. In each of these variants, we modify the category (label) of an object in the scene graph and query the SG2I model with this altered scene graph $G'$ to determine whether the removed object might inadvertently reappear in the generated output. The three query types are defined as follows:
\begin{itemize}
    \item \textbf{Fuzzy Query Attack}: Replace the unlearned object’s category label in the scene graph with a more general or ambiguous label (e.g., replacing "man" with "person").
    
    \item \textbf{Neighbor Query Attack}: Modify the category label of an object in the scene graph adjacent to, but not directly connected to, the unlearned object node.
    
    \item \textbf{Singleton Query Attack}: Modify the category label of an isolated object in the scene graph, unconnected to the unlearned object.
\end{itemize}
For this test, we specifically selected Obj-MK-PA and Obj-MK-NS, as they demonstrated the best performance in the previous experiments.

We present a visual result in Figure~\ref{fig:indirect_leakage} to illustrate the outcomes across these attacks.
The resulting generation remains consistently lacking, signifying that the object learning is robust to small alterations in the scene graphs.
In other words, these query strategies failed to bypass the unlearning process or indirectly infer the unlearned object. This finding highlights the robustness of the unlearning process: objects that have been intentionally removed remain secure, and users cannot manipulate input queries to reveal forgotten content.

This robustness suggests that, in most practical scenarios, models with unlearning applied are unlikely to expose forgotten objects, even under query-based probing. However, we suspect that more targeted and advanced in-training methods might be needed to intentionally make the model leak unlearned objects. Investigating these methods and their effects will be a key focus of future research.

\vspace{2mm}
\takeawy{Our experiments confirm the certain robustness of the unlearned models, as none of the three query-based attacks by modifying objects' labels in the input scene graph are able to reveal the unlearned object.}


\subsection{Unlearning Efficiency}
Unlearning efficiency in generative models is a critical objective, given that these models are often large and complex, making retraining or even fine-tuning a highly resource-intensive and time-consuming process. To highlight the unlearning efficiency of our proposed method, we present a run time comparison among retrain, fine-tuning-based (Obj-NG, Obj-MK-PA, and Obj-MK-NS), and model redaction-based (Obj-IF) methods. Considering the time consumption is similar for each large class of methods, we only show the large class comparisons in Table~\ref{tab:run_time}.

We can observe that, both fine-tuning and model redaction significantly improve the efficiency of unlearning in generative models, providing practical alternatives to the high computational demands of retraining. Notably, effective unlearning can be achieved through fine-tuning in as little as one-thirtieth of the time required for retraining. Moreover, model redaction, on the other hand, can complete unlearning at extreme times due to its one-time-change nature.

However, while model redaction methods stand out in efficiency, our experiments suggest that directly adapting existing redaction techniques to the object unlearning task falls short in terms of effectiveness. When considering the dual requirements of unlearning effectiveness and efficiency, fine-tuning emerges as the more suitable approach, striking a better balance between achieving the desired unlearning outcomes and minimizing computational overhead.

\vspace{2mm}
\takeawy{For object unlearning, fine-tuning methods make unlearning much more efficient while still being effective at removing specific objects. They are one of the best options for balancing effectiveness and efficiency, providing a practical and better alternative to retraining for object unlearning tasks.}

\section{Discussion}
This work presents a pioneering effort to address the complex challenges of unlearning specific entities from complex models. Despite the significant progress made, there still remains open questions for further exploration.

When compared to models built on large-scale datasets, models trained on small datasets typically have weaker generalization capabilities, which means they are more likely to experience overfitting or performance degradation after unlearning. Thus, the small model may more easily forget a specific object completely.
In contrast, large dataset models posses greater generalization, even after performing object unlearning, due to the model's deeper understanding of data patterns, it may be more difficult to completely forget a specific object, and thus our configuration above would need to be tuned further.

Our current proposal identifies a specific component of a complex architecture on which to perfrom the unlearning. However increasingly complex models may further add modules within the image generation pipeline, thereby increasing the challenge to determine which components are the most effective to modify for unlearning. This is clear in the differing principles of image decoders to GAN-based and diffusion-based models.


The integration of scene graphs provides improved practicability and extensibility for the proposed framework.
One advantage of our proposal is the enablement of unlearning multimodal data sources. In this study, we have focused on unlearning visual information through the scene graph. However, scene graphs are agnostic to the output mode, thereby making the framework generalizable to multimodal data sources. For instance, textual data (e.g., a caption) can be represented as a scene graph and thus incorporated to enhance image generation (e.g., text-to-image generation) in a multimodal manner.
Further, object unlearning can also be harnessed to remove sensitive information contained within generated text, such as sensitive entities. In this work we focus on the image synthesis model and leave text synthesis for future work.


While we have formulated new metrics to measure object-level unlearning, existing distance metrics may not fully capture the true effectiveness of unlearning, particularly when done in the interest of privacy. Consider the instance, where the metrics may suggest successful unlearning with large distances, yet the visual features of the unlearned objects remain highly recognizable. This discrepancy highlights a need for improvement developing suitable verification techniques to determine what constitutes successful object unlearning. 

Lastly, while we validated the query-based attack in this study and demonstrated that our method effectively resisted it, real-world scenarios may involve more powerful threat models. We are concerned that the masking-based unlearning method, despite its strong performance in this study, may be vulnerable to attacks exploiting differences between the model before and after unlearning \cite{DBLP:conf/ccs/Chen000HZ21}. Therefore, further research is needed to enhance the security and privacy of object unlearning methods.

\section{Conclusion}
In this paper, we introduce a novel framework for object unlearning, specifically addressing the limitations of current unlearning approaches on handling granular unlearning request. Unlike traditional sample or feature unlearning methods, our scene graph-based approach provides a targeted unlearning mechanism that selectively removes sensitive objects while preserving the utility of other, non-requested elements in the data. We validate the effectiveness of this framework through extensive evaluations on image reconstruction and synthesis tasks, demonstrating its superior ability to obscure unlearned objects without compromising the overall quality of the generated images. By leveraging influence functions to approximate the unlearning process, we mitigate the computational costs typically associated with generative models. Our findings highlight the importance of fine-grained unlearning in addressing user's varying data removal requests, while preserving the integrity and utility of the original dataset. This work lays the foundation for more precise unlearning methods and paves the way for future research aimed at enhancing privacy protections in MLaaS platforms without sacrificing model utility.

\newpage

\bibliographystyle{IEEEtran}
\bibliography{ref}



\appendix


\begin{figure*}[t]
    \centering
    \begin{subfigure}[t]{0.49\textwidth}
        \centering
        \includegraphics[width=\textwidth]{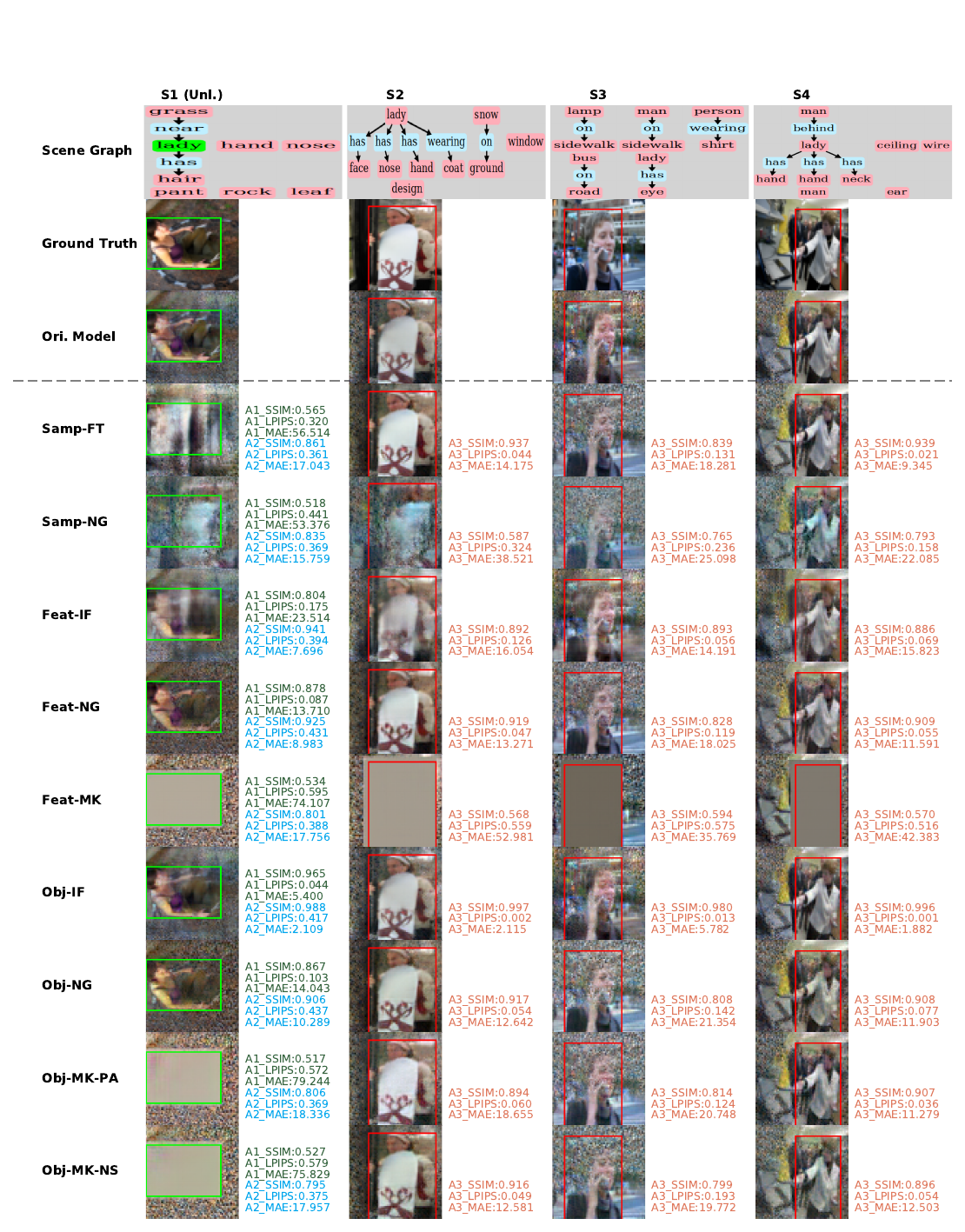}
        \caption{\textit{image reconstruction}.}
        \label{fig:exp_recon_2}
    \end{subfigure}
    \hfill
    \begin{subfigure}[t]{0.49\textwidth}
        \centering
        \includegraphics[width=\textwidth]{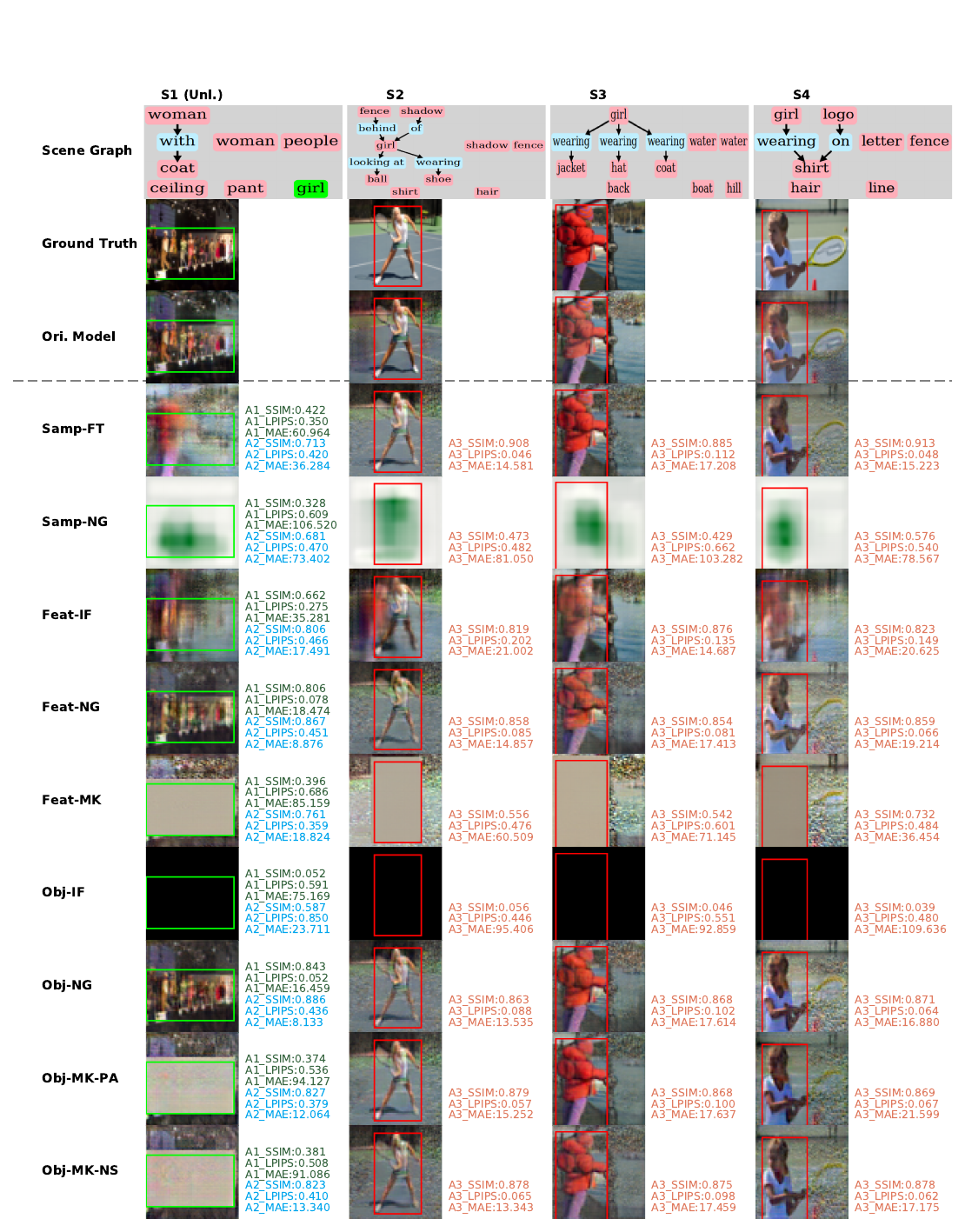}
        \caption{\textit{image synthesis}.}
        \label{fig:exp_syn_2}
\end{subfigure}
    \caption{Visualization of unlearning verification (additional results).}
    \label{fig:exp_recon_syn_2}
\end{figure*}

\begin{figure*}[t]
    \centering
    \begin{subfigure}[t]{1\textwidth}
        \centering
        \includegraphics[width=.9\textwidth]{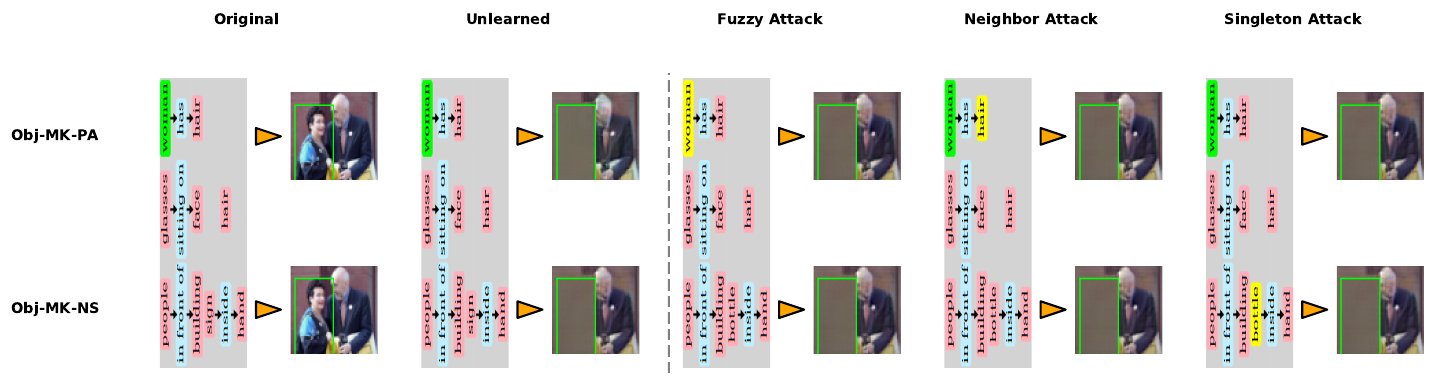}
        \caption{\textit{image reconstruction}.}
        \label{sunfig:exp_leak_2_recon}
    \end{subfigure}
    \begin{subfigure}[t]{1\textwidth}
        \centering
        \includegraphics[width=.9\textwidth]{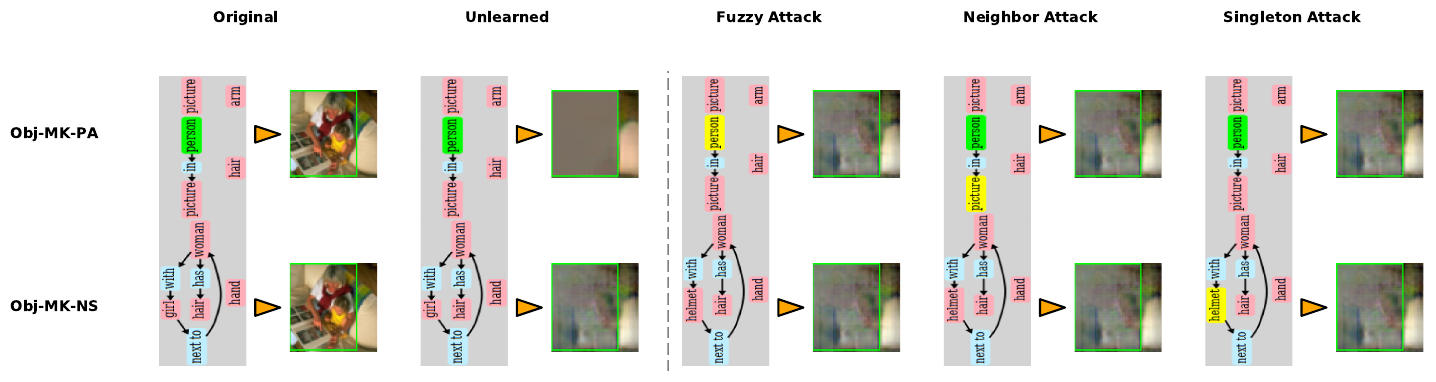}
        \caption{\textit{image synthesis}.}
        \label{subfig:exp_leak_2_synth}
\end{subfigure}
    \caption{Indirect leakage tests (additional results).}
    \label{fig:indirect_leakage_2}
\end{figure*}

\begin{figure}[!ht]
    \centering
     \includegraphics[width=\linewidth]{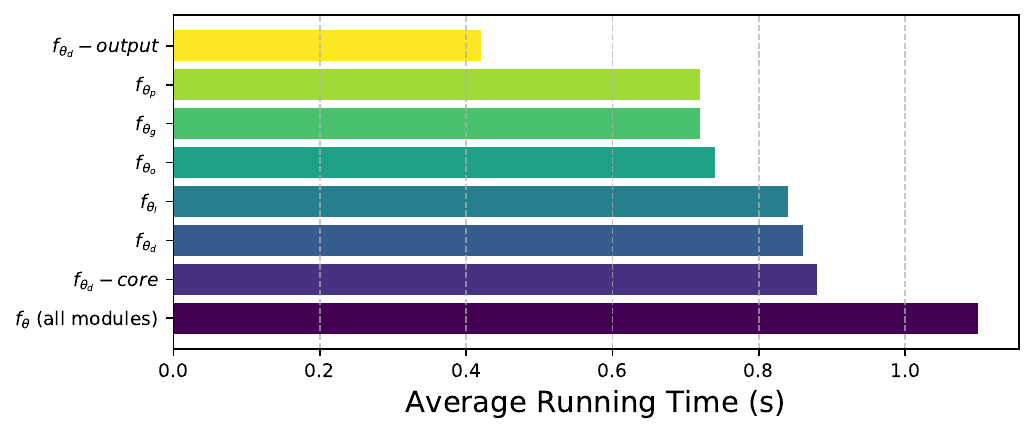} 
    \caption{Influence based unlearning run time when redaction applied on difference modules.}
    \label{fig:run_time}
\end{figure}

\begin{figure*}[t]
    \centering
     \includegraphics[width=\textwidth]{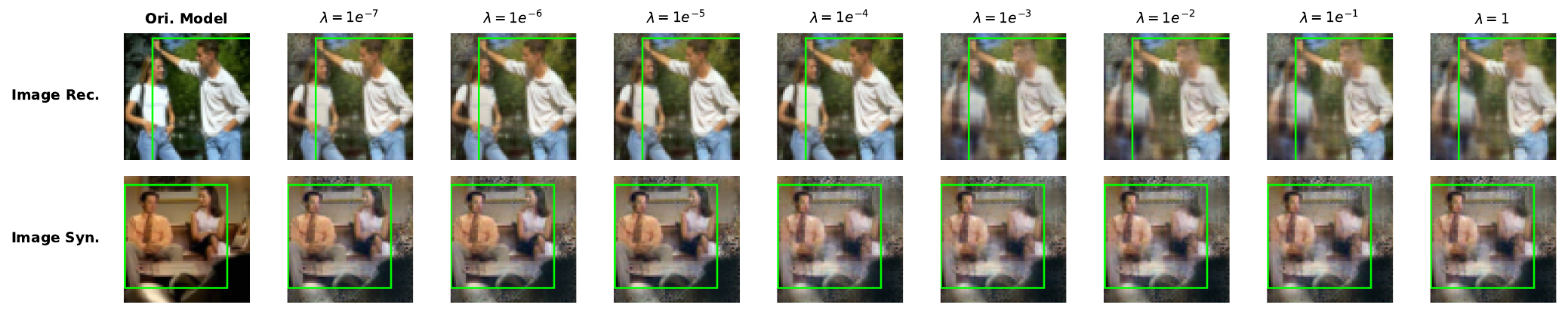}
     \vspace{-5mm}
    \caption{Visualization of ablation test on selection of different scalars $\lambda$.  }
    \label{fig:exp_abla_lambda}
\end{figure*}

\begin{figure*}[t]
    \centering
     \includegraphics[width=\textwidth]{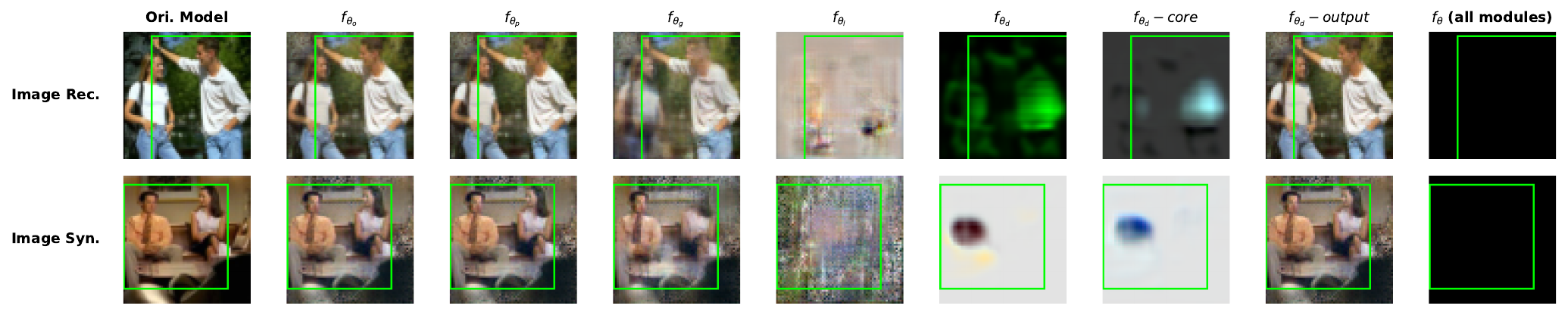}
     \vspace{-5mm}
    \caption{Visualization of the ablation test on the redaction of different modules in the SG2I model. ‘$f_{\theta_d}$-core’ represents unlearning only the core layers of the image decoder, while ‘$f_{\theta_d}$-output’ represents unlearning only the output layer of the image decoder.}
    \label{fig:exp_abla_module}
\end{figure*}

\section*{A. Details of Obj-IF Method.}
\label{app:method_mpr}

In OBJ-IF, calculating $\mathcal{L}_{\Delta \mathcal{O}}$ is challenging as we \textit{cannot} naively express it as $\mathcal{L}_{\Delta \mathcal{O}}=\sum_{I_i \in \Delta \mathcal{O}} l\left(f_{\theta^*}\left(G_i, I_i\right), I_i \right)$. In this form, it is equivalent to sample unlearning on the whole samples including the unrequested objects.
When constructing the SG2I generator, we combine the object's visual embedding ($z_v$) with its object embedding ($z_o$) and bounding box embedding ($z_b$), producing fused object embedding $z_s \mathrm{concat}(z_v, z_b, z_o)$.
To unlearn the visual feature of the object, i.e., $z_v$, from the model, we modify the fused object embedding by setting $z_v = 0$ for all objects that are to be removed. This results in $z_{s} = \mathrm{concat}(\mathbf{0} \in \mathbb{R}^{d}, z_b, z_o)$ for each $\mathbf{o} \in \Delta \mathcal{O}$. We denote this modified fused object embedding as $z_{s^{+}}$.
Then, $\mathcal{L}_{\Delta \mathcal{O}}$ can be expressed as:
\begin{equation}
\begin{aligned}
     \mathcal{L}_{\Delta \mathcal{O}} & =\sum_{I_i  \in \Delta \mathcal{O}} l\left((f_{\theta_g^*} \circ f_{\theta_l^*} \circ  f_{\theta_d^*}) (z_{s^+}) , I_i \right) \\
     z_{s^+} &= \begin{cases} \mathrm{concat}(\mathbf{0} \in \mathbb{R}^{d}, z_b, z_o), & \mathbf{o} \in \Delta \mathcal{O} \\ 
\mathrm{concat}(z_v, z_b, z_o), & \text { other nodes }\end{cases}
\end{aligned}
\end{equation}

The estimated parameter change can be derived as:
\begin{equation}
\begin{aligned}
\Delta \theta= & H_{\theta^*}^{-1} \underbrace{\sum_{I_i  \in \Delta \mathcal{O}} \nabla_{\theta^*} l_{\Delta\mathcal{O}}((f_{\theta_g^*} \circ f_{\theta_l^*} \circ  f_{\theta_d^*}) (z_{s^+}) , I_i ))}_{\text{unlearned objects}} \\
& - H_{\theta^*}^{-1} \sum_{I_i  \in \Delta \mathcal{O}} \nabla_{\theta^*} l_{\Delta\mathcal{O}}((f_{\theta_g^*} \circ f_{\theta_l^*} \circ  f_{\theta_d^*}) (z_{s}) , I_i )), 
\end{aligned}
\end{equation}
where $l_{\Delta\mathcal{O}}(I', I) \triangleq l(I'[\Delta\mathcal{O}], I[\Delta\mathcal{O}]) $  measures the discrepancy or difference between the region of removed object in the generated image $I'$ and the region of object in the original image $I$. 
Given that directly calculating the inverse Hessian matrix is computationally expensive, we can instead use fast Hessian-vector products (HVPs) \cite{DBLP:journals/neco/Pearlmutter94} to speed up the process reducing computational complexity from $O\left(|\theta|^3+n|\theta|^2\right)$ to $O\left(n|\theta|\right)$.

The SG2I model is inherently complex, and applying model redaction on the entire model could lead to catastrophic unlearning or large computational overheads. Therefore, we propose a partial redaction approach, where only specific components of the image generator are modified, rather than the entire model. 

The primary challenge in this approach is identifying which module is most responsible for the removal of the requested object’s visual information while ensuring the overall scene semantics and structure are preserved. 
To address this, we target the graph representation learner (GRL) $f_{\theta_g}$ for mode redaction, instead of the entire model. This decision is informed by the following considerations:

First, we formulate unlearning as a problem of node feature unlearning within a graph, and utilize an influence function-based parameter estimation technique. This method computes the influence of masking the visual features embedded in $z_v$ which is part of the concatenated object embedding $z_s = \mathrm{concat}(z_v, z_b, z_o)$. 
The GRL is primarily responsible for processing these embedded features of different objects and mapping them to the layout and image generation stages, making it the optimal target for unlearning without disrupting the entire generative process.

Second, when comparing the properties of different modules, the visual extractor encodes the majority of the model’s knowledge regarding visual features, typically through complex, pretrained models such as Vision Transformers (ViT), making it difficult and inefficient to modify. Modifying the image decoder could degrade the overall image quality, while modifying the layout predictor may result in incorrect object placement or overlap. Predicate embedders do not store detailed object-specific information. Thus, the GRL, which maps visual, object, and predicate information into layout and image generation, is the most suitable target for modification in our redaction approach.


\section*{Additional Experimental Results of Visualization} \label{app:additional}
In this section, we provide some additional results of visualizations. These results are consistent with the results presented in the main body. The image generation task involves a high degree of randomness, and due to space limitations, we cannot display all the results here. However, we will include as many results as possible in the open-source repository.

\section*{Ablation Test of Obj-IF Method} \label{app:exp_abla}
\paragraph{Influence of Redaction on Different Modules}
In Section~\ref{app:method_mpr} we propose a partial model redaction strategy to modify the model given the parameter estimation. It is necessary to explore the influence of redaction on different modules of the SG2I Model to unlearned model’s performance.

The results shown in Figure~\ref{fig:exp_abla_module} indicate that modifying different modules in the model significantly affects its unlearning performance. Modifying all modules leads to catastrophic failure, as seen in the final column where the generated images become completely unrecognizable, rendering the model ineffective. Modifying the decoder also severely distorts the generated images, demonstrating its crucial role in preserving output quality. In contrast, modifying the object or predicate embedders has minimal effect, as the images retain much of their original feature.
Our proposed method, which modifies only the graph representation learner (GRL), achieves the most balanced and effective unlearning. By targeting the GRL, the unlearning process effectively removes the requested object without introducing excessive distortions or impairing the overall image quality. This ensures that the visual utility of the model is retained, making it the most suitable and effective method for object unlearning purpose.

\paragraph{Influence of Redaction with Different Scalars}
In this experiment, we explore the Influence of Redaction with Different Scalars to determine how varying the degree of information removal affects the performance and stability of the model. 

The results shown in Figure~\ref{fig:exp_abla_lambda} demonstrate the impact of varying scalar multiplier $\lambda$ on the unlearning performance of O-Unl. As $\lambda$ increases from $1e^{-7}$ to $1$, the unlearning effect becomes more pronounced. For smaller values of $\lambda$ (e.g., $1e^{-7}$  to $1e^{-5}$ ), the unlearned object remains relatively clear in both image reconstruction and image synthesis, indicating a weaker unlearning effect. As $\lambda$ grows larger, the unlearned object becomes increasingly blurred or indistinct, particularly noticeable in the reconstructed images where facial features become unrecognizable at $1e^{-3}$  and beyond. This suggests that larger $\lambda$ values correspond to more effective unlearning.
However, tuning $\lambda$ too aggressively (e.g., at $1$) can introduce excessive blurring and distortion, not only to the unlearned object but also to the surrounding features. A balanced choice of $\lambda$, such as in the mid-range values (e.g., $1e^{-3}$), allows for sufficient unlearning while preserving the quality of the remaining features in the scene.
Moreover, our offline tests indicate that models pretrained at different levels exhibit varying sensitivity to parameter changes. Developing a stable and consistent solution to address this sensitivity remains an area for future research.

Additionally, within the Obj-IF, we compare the efficiency of applying model redaction across different modules. From the results shown in Figure~\ref{fig:run_time}, it is evident that redaction across all modules is time-consuming. However, the proposed partial module redaction strategy significantly improves unlearning efficiency by enabling selective redaction. Furthermore, the selection of redaction on the graph representation learner demonstrates a reasonable running time, reinforcing the practicality of our approach.



%



\end{document}